\documentclass{article}

\PassOptionsToPackage{numbers, compress}{natbib}

 \usepackage[preprint]{neurips_2025}

\usepackage[utf8]{inputenc} 
\usepackage[T1]{fontenc}    
\usepackage{hyperref}       
\usepackage{url}            
\usepackage{booktabs}       
\usepackage{amsfonts}       
\usepackage{nicefrac}       
\usepackage{microtype}      
\usepackage{xcolor}         
\usepackage{caption}
\usepackage{pifont} 

\usepackage{graphicx}
\usepackage{multirow}
\usepackage[table]{xcolor}
\usepackage{tabularx} 
\usepackage{makecell} 
\usepackage{amsmath}
\usepackage{bbm} 
\usepackage{booktabs}
\usepackage{multirow}
\usepackage{subcaption}
\usepackage{wrapfig}

\definecolor{best}{RGB}{255, 204, 204}    
\definecolor{second}{RGB}{255, 229, 204}  
\definecolor{third}{RGB}{255, 255, 204}   
\newcommand{\bestc}[1]{\cellcolor{best}{#1}}
\newcommand{\secondc}[1]{\cellcolor{second}{#1}}
\newcommand{\thirdc}[1]{\cellcolor{third}{#1}}

\title{\textsc{OneWorld:} Taming Scene Generation with \\ 3D Unified Representation Autoencoder}

\author{
Sensen Gao$^{1*}$, Zhaoqing Wang$^{2*}$, Qihang Cao$^{3}$, Dongdong Yu$^{2}$, Changhu Wang$^{2}$\\
\textbf{Tongliang Liu}$^{4,1\dagger}$, \textbf{Mingming Gong}$^{5,1\dagger}$, \textbf{Jiawang Bian}$^{6\dagger}$  \\
  $^{1}$ Mohamed bin Zayed University of Artificial Intelligence \quad 
  $^{2}$ AISphere  \\
  $^{3}$ Shanghai Jiao Tong University \quad $^{4}$ University of Syndey \\
  $^{5}$ University of Melbourne \quad $^{6}$ Nanyang Technological University \\
  * Co-first authors. \quad $^{\dagger}$Corresponding authors.\\
}

\begin{document}
\maketitle

\vspace{-10mm}
\begin{figure*}[h]
    \centering
    \includegraphics[width=\textwidth]{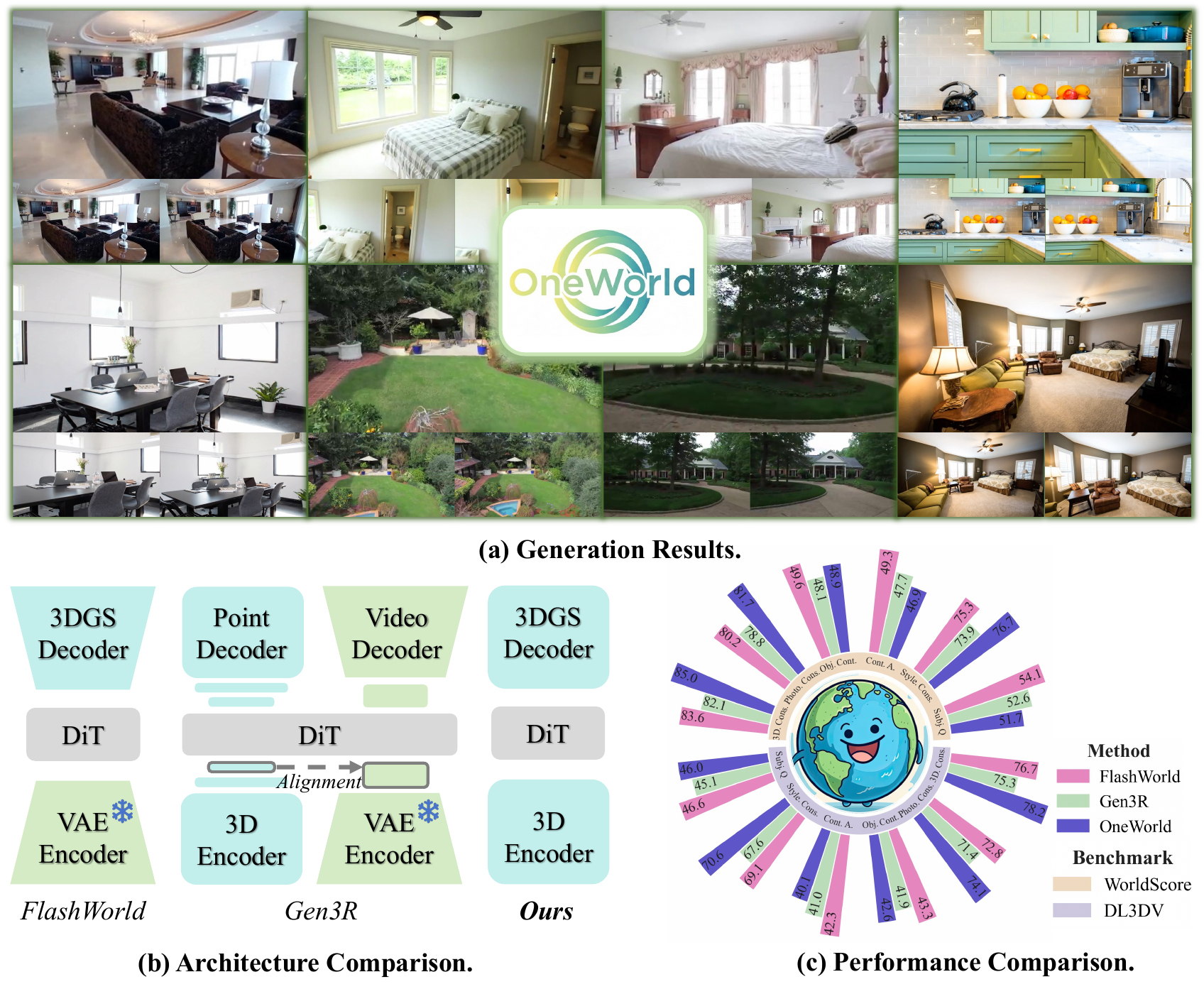}
    \vspace{-7mm}
    \caption{(a) OneWorld generates 3DGS from a single view and renders novel views. (b) Architecture: FlashWorld~\cite{li2025flashworld} diffuses in compressed video latents; Gen3R~\cite{huang2026gen3r} compresses 3D features to align a 3D foundation encoder with video latents but can only generate geometry and appearance separately, rather than jointly. OneWorld generates directly in a unified 3D representation space, without compression or separate generation. (c) Performance comparison on WorldScore~\cite{duan2025worldscore} and DL3DV~\cite{ling2024dl3dv}.}
    \label{fig:teaser}
    \vspace{-5mm}  
\end{figure*}


\begin{abstract}
Existing diffusion-based 3D scene generation methods primarily operate in 2D image/video latent spaces, which makes maintaining cross-view appearance and geometric consistency inherently challenging. To bridge this gap, we present OneWorld, a framework that performs diffusion directly within a coherent 3D representation space. Central to our approach is the 3D Unified Representation Autoencoder (3D-URAE); it leverages pretrained 3D foundation models and augments their geometry-centric nature by injecting appearance and distilling semantics into a unified 3D latent space. Furthermore, we introduce token-level Cross-View-Correspondence (CVC) consistency loss to explicitly enforce structural alignment across views, and propose Manifold-Drift Forcing (MDF) to mitigate train–inference exposure bias and shape a robust 3D manifold by mixing drifted and original representations. Comprehensive experiments demonstrate that OneWorld generates high-quality 3D scenes with superior cross-view consistency compared to state-of-the-art 2D-based methods. Our code will be available at \href{https://github.com/SensenGao/OneWorld}{https://github.com/SensenGao/OneWorld}.
\end{abstract}

\section{Introduction}
3D scene generation maps conditioning inputs, such as text or images, to a 3D scene representation, producing either explicit outputs like point clouds or 3DGS~\cite{kerbl20233d}, or implicit outputs like video. It has emerged as vital technology for gaming, robotics, and VR/AR by enabling the creation of photorealistic environments at scale~\cite{ding2025understanding,kong20253d,ye2026world,zhang2025advances}.
This capability provides essential training data for simulations and offers powerful new tools for creative content design.
By jointly modeling scene geometry, appearance, and spatial structure, it serves as the backbone for building interactive and geometrically consistent digital worlds.

Early 3D scene generation methods often rely on pretrained 2D generative priors, either through optimization-based score distillation~\cite{lin2023magic3d,poole2022dreamfusion,tang2023dreamgaussian,wang2023prolificdreamer} or by synthesizing multi-view images or videos followed by 3D reconstruction~\cite{gao2024cat3d,hao2025gaussvideodreamer,liu2023syncdreamer,shi2023mvdream,wu2024reconfusion}. While these approaches can produce plausible results, they are limited by expensive per-scene optimization and inconsistent geometry due to weak explicit 3D modeling. More recently, several works move generation into multi-view or video latent spaces by freezing an image or video VAE (See Fig.~\ref{fig:teaser} (b)) and learning a latent-to-3D decoder (often 3DGS~\cite{kerbl20233d}) while adapting the diffusion backbone~\cite{yang2025prometheus,go2025splatflow,go2025videorfsplat,li2025flashworld,dai2025fantasyworld,go2025vist3a}. This improves efficiency, but 2D latent spaces provide limited explicit 3D coupling across views, which may weaken multi-view coherence.
Meanwhile, transformer-based 3D feed-forward models~\cite{wang2025vggt,wang2024dust3r,wang2025pi,cabon2025must3r,lin2025depth,leroy2024grounding} have advanced rapidly. Models such as Dust3R~\cite{wang2024dust3r}, VGGT~\cite{wang2025vggt} and $\pi^3$~\cite{wang2025pi} achieve strong performance in a wide range of 3D downstream tasks. Inspired by recent image generation approaches that operate directly in the representation space of vision foundation models (\textit{e.g.}, DINOv2~\cite{oquab2024dinov2}) rather than relying on VAEs~\cite{zheng2025diffusion,chen2025aligning,shi2025latent,bi2025vision}, we explore whether it is possible to learn 3D scene generation directly within the representation space of a 3D foundation model.

The concurrent work Gen3R~\cite{huang2026gen3r} compresses the 3D representation space to align with a VAE video latent space~\cite{wan2025wan}, limiting the representational capacity of the learned 3D representation. In addition, it generates geometry (point clouds) and appearance (video) separately (See Fig.~\ref{fig:teaser} (b)), constraining a unified 3D representation with coherent geometry and high-fidelity appearance. Furthermore, its reliance on video-latent alignment constrains geometric modeling to small-baseline view variations, limiting generalization to large viewpoint changes.

To address these challenges, we present OneWorld, a framework that performs diffusion within a coherent 3D representation space, enabling sampling within a coherent world rather than per-view image or video latents.
Central to our approach is the 3D Unified Representation Autoencoder (3D-URAE); it leverages pretrained 3D foundation models and augments their geometry-centric nature by injecting appearance and distilling semantics into a unified 3D latent space.
The appearance-injection branch complements semantic tokens with appearance tokens, allowing unified 3D tokens to be decoded into a renderable 3DGS representation with consistent visual identity. The semantic-distillation branch transfers knowledge from a vision foundation model (VFM) to regularize the 3D tokens onto a compact, semantically meaningful manifold, reducing learning complexity and improving diffusion efficiency in 3D representation space.

Beyond the unified representation, we explicitly preserve cross-view structure during diffusion training by introducing a token-level Cross-View-Correspondence (CVC) consistency loss. Specifically, we enforce the target-view tokens to retain the correspondence patterns induced by the conditioning-view tokens, so that denoising preserves structural consistency rather than merely minimizing the mean error between 3D representations. Moreover, we identify sampling drift induced by train--inference exposure bias as a key factor that degrades 3D generation quality: at inference the model must denoise its own intermediate predictions, so small errors accumulate over steps, and this effect is amplified in 3D by coupled cross-view constraints.
To mitigate this issue, we propose Manifold-Drift Forcing (MDF), which trains the 3DGS decoder with a mixture of diffusion-sampled (drifted, off-manifold) and original (on-manifold) unified representations during decoding, shaping a robust 3D representation manifold for stable diffusion sampling, and improving appearance consistency and cross-view coherence. Extensive experiments on RealEstate10K~\cite{zhou2018stereo}, DL3DV~\cite{ling2024dl3dv}, and WorldScore~\cite{duan2025worldscore} demonstrate that OneWorld generates high-quality 3D scenes with strong cross-view consistency, significantly outperforming prior 2D-based approaches.

Our main contributions are summarized as follows:
\begin{itemize}
    \item We propose \textbf{OneWorld}, a diffusion-based 3D scene generation framework that operates in a 3D feature space built upon pretrained 3D foundation models and unified by our 3D Unified Representation Autoencoder (3D-URAE), which injects appearance and distills semantics to jointly encode geometry, appearance, and semantics.
    \item We introduce a cross-view correspondence-preserving loss for conditional diffusion training in unified 3D space, explicitly enforcing structural correspondence between the target view and conditioning views to improve cross-view consistency.
    \item We identify sampling drift from train–inference mismatch in 3D generation and propose Manifold-Drift Forcing, which trains the 3DGS decoder on mixed diffusion-sampled and ground-truth latents to shape a robust 3D manifold for stable sampling, improving appearance consistency and cross-view coherence.
\end{itemize}

\section{Related Work}
\subsection{3D Scene Generation}
\textbf{Diffusion-based Iterative 3D Scene Generation.} 
Early 3D generation methods typically leverage pretrained 2D generative models~\cite{rombach2022high,podell2023sdxl,zhang2023adding,peebles2023scalable} to provide strong generative priors. One line of work employs Score Distillation Sampling (SDS)~\cite{lin2023magic3d,poole2022dreamfusion,tang2023dreamgaussian,wang2023prolificdreamer} to optimize a 3D representation, such as 3DGS~\cite{kerbl20233d} or NeRF~\cite{mildenhall2021nerf}, by aligning rendered views with the distribution of a pretrained 2D diffusion model. Although these iterative generation approaches achieve notable progress, they often suffer from cross-view semantic inconsistency due to the lack of explicit multi-view constraints. Moreover, generating each individual 3D scene requires extensive iterative optimization, leading to substantial computational cost and limited scalability.

\textbf{Multi-View Reconstruction-Based 3D Scene Generation.}
Another line of methods first synthesizes multi-view images or videos using pretrained 2D diffusion models, followed by 3D reconstruction through multi-view synthesis~\cite{chen2024mvsplat360,gao2024cat3d,hao2025gaussvideodreamer,liu2024reconx,liu2023syncdreamer,sargent2023zeronvs,shi2023mvdream,sun2024dimensionx,wu2024reconfusion,zhao2024genxd} or incremental outpainting~\cite{chung2023luciddreamer,fridman2023scenescape,schwarz2025recipe,yu2024wonderjourney,yu2025wonderworld}. Compared to iterative optimization-based approaches, these methods significantly improve generation efficiency by avoiding per-scene optimization. However, despite this speed advantage, they still lack explicit 3D reasoning, as the underlying generative process is grounded in 2D RGB priors. As a result, they often produce inconsistent geometry and weak multi-view fidelity.

\textbf{3D Scene Generation in 2D Latent Spaces.}
The difference between the 3D generation method introduced here and Multi-View Reconstruction-Based methods is that one uses a Diffusion model to generate multi-view images or videos, while this method generates multi-view latents~\cite{yang2025prometheus,go2025splatflow,li2024director3d} or video latents~\cite{li2025flashworld,dai2025fantasyworld,go2025videorfsplat,go2025vist3a}. The original image VAE or video VAE encoder is frozen, and a decoder for decoding 3D representations is trained, usually implemented as 3DGS. Early approaches, such as Prometheus~\cite{yang2025prometheus} and SplatFlow~\cite{go2025splatflow}, adapt SD-VAE~\cite{podell2023sdxl} to reconstruct 3DGS representations from multi-view latents, and subsequently fine-tune the image generation model. Building upon this latent-to-3D reconstruction paradigm, later works including FlashWorld~\cite{li2025flashworld}, VIST3A~\cite{go2025vist3a}, and FantasyWorld~\cite{dai2025fantasyworld} replace SD-VAE~\cite{podell2023sdxl} with Wan-VAE~\cite{wan2025wan}, enabling reconstruction from video latents and extending the framework to video generation models. Learning in the multi-view latent space offers improved computational efficiency and better compatibility with pretrained diffusion backbones compared to directly modeling multi-view image generation; however, since 2D image or video latent spaces do not explicitly enforce cross-view identity consistency, this formulation often results in geometric inconsistency.

\subsection{Representation Autoencoder}
A common line of work in image and video generation uses VAEs~\cite{kingma2013auto,esser2024scaling,van2017neural} to compress data into low-dimensional latent spaces, facilitating the training of latent diffusion models~\cite{rombach2022high}. However, such compression inevitably results in information loss. Recently, some studies~\cite{zheng2025diffusion,shi2025latent,chen2025aligning,bi2025vision} have explored a different route: employing a frozen, pretrained representation encoder and training only the decoder to reconstruct images from high-dimensional semantic features. Known as Representation Autoencoders (RAEs), this approach demonstrates that training diffusion transformers~\cite{peebles2023scalable} in such latent spaces achieves faster convergence and better performance than VAEs. Motivated by this shift, we design a 3D Unified RAE on top of a pretrained 3D foundation model so diffusion happens directly in a geometry-aware representation space.

\begin{figure*}[t]
    \centering
    \includegraphics[width=\textwidth]{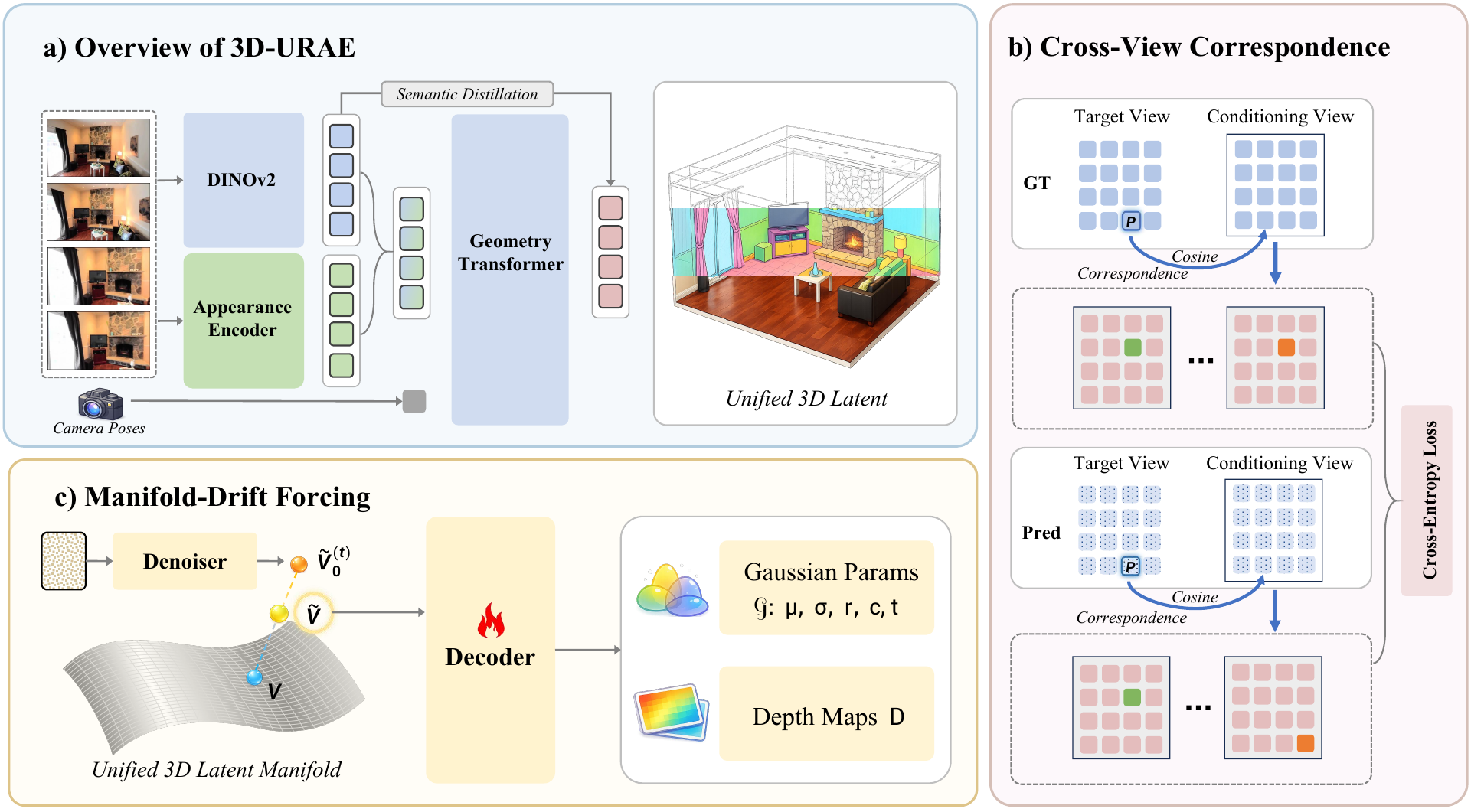}
    \caption{\textbf{Overview of the proposed OneWorld framework.}
\textbf{(a)} We construct a \textbf{unified 3D representation space} by introducing \textbf{appearance injection} and \textbf{semantic distillation}.
\textbf{(b)} During DiT~\cite{peebles2023scalable} training, we incorporate \textbf{cross-view correspondence}, preserving cross-view geometric token correspondences from the target view to the conditioned view.
\textbf{(c)} \textbf{Manifold-drift forcing:} we augment the original 3D manifold by mixing ground-truth 3D features with sampled 3D features, enabling a more robust 3D decoder.
    }
    \label{fig:method}
    \vspace{-0.5em}
\end{figure*}

\section{Method}
\label{sec:method}

We present OneWorld, a diffusion framework that performs scene generation directly in a unified, geometry-aware 3D representation space (See Fig.~\ref{fig:method}). First, we build a 3D Unified Representation Autoencoder (3D-URAE) on top of a feed-forward 3D foundation model by injecting appearance cues and distilling semantic structure into geometry tokens, producing renderable and semantically organized 3D latents (Sec.~\ref{sec:rae}). Next, we train a conditional diffusion model in this unified space and introduce a cross-view correspondence preservation regularizer to maintain consistent token-level correspondences between the target and conditioning views, improving cross-view structural coherence (Sec.~\ref{sec:diffusion_ctcp}). Finally, to address train--inference exposure bias that causes diffusion samples to drift off the 3D-URAE manifold, we propose manifold-drift forcing, which trains the downstream 3D decoder on interpolations of ground-truth and sampled latents to improve robustness and stabilize multi-view sampling (Sec.~\ref{sec:mdf}).

\subsection{3D Unified Representation Autoencoder}
\label{sec:rae}

\textbf{Preliminary.}
We adopt a 3D foundation model for feed-forward scene reconstruction. In this work, the instantiated model is $\pi^3$~\cite{wang2025pi}, denoted as $\mathcal{F}$, which infers multiple key 3D quantities of a scene from observed views together with their camera parameters. Given $N$ input images $\mathcal{I} \in \mathbb{R}^{N \times H \times W \times 3}$, $\pi^3$ first leverages a vision foundation model (VFM) (\textit{e.g.},  DINOv2~\cite{oquab2024dinov2}) as an image patchifier or tokenizer, denoted as $\mathcal{E}_{\mathrm{patch}}$, to obtain per-view visual tokens:
\begin{equation}
\mathcal{E}_{\mathrm{patch}}:\ \mathcal{I} \rightarrow \mathcal{Z} \in \mathbb{R}^{N \times h \times w \times C},
\end{equation}
where $h \times w$ is the token grid resolution and $C$ is the token channel dimension. To enable camera-controllable generation and to directly use the dataset-provided ground-truth camera coordinate system without additional conversions, $\pi^3$ takes the per-view camera parameters $\mathcal{T} \in \mathbb{R}^{N \times C_T}$ (\textit{e.g.}, intrinsics and extrinsics in the dataset convention) as explicit inputs. The geometry encoder $\mathcal{E}_{\mathcal{V}}$ then jointly encodes the patchified tokens and camera parameters into geometry tokens:
\begin{equation}
\mathcal{E}_{\mathcal{V}}:\ (\mathcal{Z}, \mathcal{T}) \rightarrow \mathcal{V} \in \mathbb{R}^{N \times h_v \times w_v \times C_v},
\end{equation}
where $h_v \times w_v$ is the spatial resolution of geometry tokens and $C_v$ is the corresponding feature dimension. Finally, the geometry tokens are decoded by prediction heads $\mathcal{D}_{\mathcal{V}}$ into a 3D Gaussian Splatting representation and depth maps:
\begin{equation}
\mathcal{D}_{\mathcal{V}}:\ \mathcal{V} \rightarrow (\mathcal{G}, \mathcal{D}),\qquad
\mathcal{G} \in \mathbb{R}^{N \times H \times W \times C_{\mathrm{GS}}},\ 
\mathcal{D} \in \mathbb{R}^{N \times H \times W \times 1},
\end{equation}
where $C_{\mathrm{GS}}$ denotes the dimensionality of the 3DGS parameterization used in our implementation.

\begin{figure*}[t]
    \centering
    \includegraphics[width=0.9\textwidth]{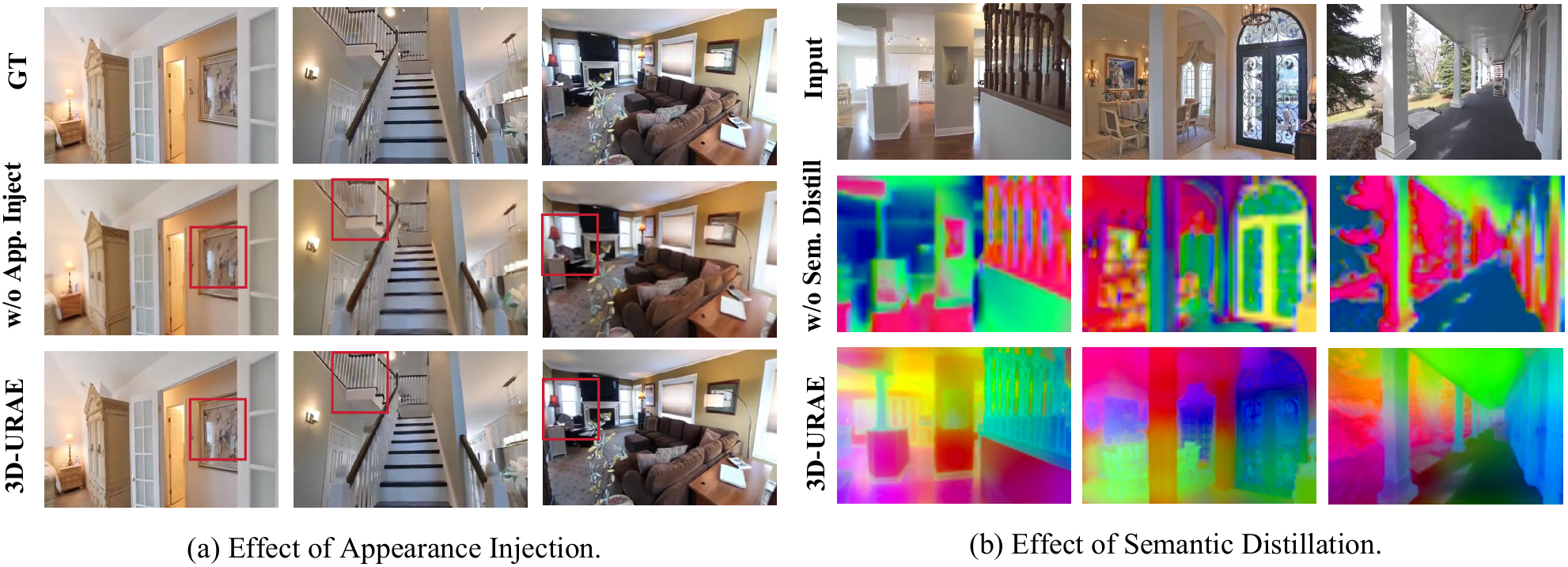}
    \vspace{-2mm}
    \caption{\textbf{Visualization of the effects of appearance injection and semantic distillation.}
We visualize the reconstruction results with and without \textbf{appearance injection} and present feature visualizations with and without \textbf{semantic distillation}.
    }
    \label{fig:ablation_recon}
    \vspace{-3mm}
\end{figure*}

\textbf{Appearance Injection Branch.}
The 3D foundation model $\mathcal{F}$ typically adopts a vision foundation model $\mathcal{E}_{\mathrm{patch}}$ (\textit{e.g.}, DINOv2) to patchify $\mathcal{I}$ into semantic tokens $\mathcal{Z}$, which may discard fine-grained appearance details due to heavy semantic abstraction. More detailed results are shown in Tab.~\ref{tab:ablation_urae_app} and Fig.~\ref{fig:ablation_recon} (a). To compensate for this loss, we introduce an appearance injection branch that extracts appearance-preserving tokens from the original images and augments the conditioning of the geometry encoder. Specifically, we implement a lightweight convolutional encoder $\mathcal{E}_{\mathrm{app}}$ to map $\mathcal{I}$ into appearance tokens $\mathcal{Z}_{\mathrm{app}} \in \mathbb{R}^{N \times h \times w \times C}$, aligned with $\mathcal{Z}$ in both resolution and channel dimension. We then concatenate $\mathcal{Z}$ and $\mathcal{Z}_{\mathrm{app}}$ along the channel axis and feed the augmented tokens with camera parameters $\mathcal{T}$ into $\mathcal{E}_{\mathcal{V}}$, yielding:
\begin{equation}
\mathcal{E}_{\mathcal{V}}:\ \big([\mathcal{Z}\,\|\,\mathcal{Z}_{\mathrm{app}}],\,\mathcal{T}\big)\ \rightarrow\ \mathcal{V} \in \mathbb{R}^{N \times h_v \times w_v \times C_v}.
\end{equation}

\begin{table*}[t]
\centering
\caption{Ablation analysis of \textbf{appearance injection} and \textbf{semantic distillation}.
\textbf{(a)} We analyze the effect of \textbf{appearance injection} from a reconstruction perspective.
\textbf{(b)} We examine the benefit of \textbf{semantic distillation} for training the generative model.}
\setlength{\tabcolsep}{4pt}

\begin{minipage}[t]{0.49\textwidth}
\vspace{-1mm}
\subcaption{Ablation on Appearance Injection Branch.}
\label{tab:ablation_urae_app}
\resizebox{\textwidth}{!}{
\setlength{\tabcolsep}{10pt}
\begin{tabular}{l|ccc}
\toprule
Method & PSNR $\uparrow$ & SSIM $\uparrow$ & LPIPS $\downarrow$ \\
\midrule
w/o App. Inject & 21.14 & 0.669 & 0.293 \\
3D-URAE      & 28.19 & 0.932 & 0.102 \\
\bottomrule
\end{tabular}}
\end{minipage}
\hfill
\begin{minipage}[t]{0.49\textwidth}
\vspace{-1mm}
\subcaption{Ablation on Semantic Distillation Branch.}
\label{tab:ablation_urae_sem}
\resizebox{\textwidth}{!}{
\setlength{\tabcolsep}{10pt}
\begin{tabular}{l|ccc}
\toprule
Method & PSNR $\uparrow$ & SSIM $\uparrow$ & LPIPS $\downarrow$ \\
\midrule
w/o Sem. Distill & 17.45 & 0.644 & 0.352 \\
OneWorld       & 21.57 & 0.735 & 0.231 \\
\bottomrule
\end{tabular}}
\end{minipage}
\vspace{-3mm}
\end{table*}
\textbf{Semantic Distillation Branch.}
Although $\pi^3$ (and similar feed-forward reconstructors such as VGGT~\cite{wang2025vggt}) uses VFM semantic tokens (\textit{e.g.}, DINOv2) as inputs, the pure 3D reconstruction supervision often yields geometry tokens $\mathcal{V}$ that are geometry-dominant and weak in semantic structure (See Fig.~\ref{fig:ablation_recon} (b)). To obtain a more compact and semantically organized 3D latent manifold that is easier for diffusion to model (More detailed results are shown in Tab.~\ref{tab:ablation_urae_sem}), we introduce a semantic distillation branch to inject semantics into $\mathcal{V}$. Specifically, we distill from the original VFM tokens $\mathcal{Z}=\mathcal{E}_{\mathrm{patch}}(\mathcal{I})$ and align them with geometry tokens $\mathcal{V}\in\mathbb{R}^{N\times h_v\times w_v\times C_v}$. A lightweight adapter $\mathcal{A}_{\mathrm{sem}}$ maps $\mathcal{Z}$ into the geometry-token space, \textit{i.e.}, $\mathcal{Z}_{\mathrm{sem}}=\mathcal{A}_{\mathrm{sem}}(\mathcal{Z})\in\mathbb{R}^{N\times h_v\times w_v\times C_v}$. Following VA-VAE~\cite{yao2025vavae}, we employ a marginal cosine similarity loss and a marginal distance matrix similarity loss; both are computed per view and then averaged over the $N$ views,
\begin{equation}
\mathcal{L}_{\mathrm{mcos}}
=\frac{1}{N}\sum_{n=1}^{N}\frac{1}{h_v w_v}\sum_{i=1}^{h_v}\sum_{j=1}^{w_v}
\mathrm{ReLU}\!\left(
1-m_1-\frac{\mathbf{z}_{n,i,j}\cdot \mathbf{v}_{n,i,j}}{\lVert\mathbf{z}_{n,i,j}\rVert\,\lVert\mathbf{v}_{n,i,j}\rVert}
\right),
\end{equation}
where $\mathbf{z}_{n,i,j}$ and $\mathbf{v}_{n,i,j}$ denote the feature vectors from $\mathcal{Z}_{\mathrm{sem}}$ and $\mathcal{V}$ at location $(i,j)$ for the $n$-th view, respectively. Moreover, letting $N_p=h_v w_v$ and flattening the spatial grid into indices $p,q\in\{1,\dots,N_p\}$, we define the marginal distance matrix similarity loss (also per view, then averaged) as
\begin{equation}
\mathcal{L}_{\mathrm{mdms}}
=\frac{1}{N}\sum_{n=1}^{N}\frac{1}{N_p^{2}}\sum_{p,q}
\mathrm{ReLU}\!\left(
\left|
\frac{\mathbf{z}_{n,p}\cdot \mathbf{z}_{n,q}}{\lVert\mathbf{z}_{n,p}\rVert\,\lVert\mathbf{z}_{n,q}\rVert}
-\frac{\mathbf{v}_{n,p}\cdot \mathbf{v}_{n,q}}{\lVert\mathbf{v}_{n,p}\rVert\,\lVert\mathbf{v}_{n,q}\rVert}
\right|
-m_2
\right).
\end{equation}
Finally, we combine the two terms as $\mathcal{L}_{\mathrm{sem}}=\mathcal{L}_{\mathrm{mcos}}+\lambda_{\mathrm{mdms}}\mathcal{L}_{\mathrm{mdms}}$.

\textbf{Training Objective for 3D-URAE.}
With the appearance injection branch and the semantic distillation branch, the 3D foundation model is transformed into 3D-URAE, and the resulting tokens $\mathcal{V}$ form a unified 3D representation that encodes geometry together with appearance and semantics. We train 3D-URAE with a differentiable 3DGS rendering loss and the semantic distillation loss; for each scene we predict $(\mathcal{G},\mathcal{D})$ and render $N_{\mathrm{novel}}$ novel views using their camera parameters, where $\mathcal{D}$ is projected to modulate the opacity of $\mathcal{G}$,
\begin{equation}
\mathcal{L}_{\mathrm{render}}
=\frac{1}{N_{\mathrm{novel}}}\sum_{n=1}^{N_{\mathrm{novel}}}
\Big(
\lVert \hat{\mathcal{I}}_n-\mathcal{I}^{\mathrm{gt}}_n\rVert_2^2
+\lambda_{\mathrm{lpips}}\ \mathrm{LPIPS}(\hat{\mathcal{I}}_n,\mathcal{I}^{\mathrm{gt}}_n)
\Big).
\end{equation}
We optimize the overall objective $\mathcal{L}_{\mathrm{URAE}}=\mathcal{L}_{\mathrm{render}}+\lambda_{\mathrm{sem}}\,\mathcal{L}_{\mathrm{sem}}$.

\subsection{Cross-view Correspondence}
\label{sec:diffusion_ctcp}
We train a conditional diffusion model on the unified 3D tokens produced by 3D-URAE. For each scene, we sample a target view token grid $\mathcal{V}^{(\mathrm{tgt})}\in\mathbb{R}^{h_v\times w_v\times C_v}$ and flatten it as $\mathbf{x}_0\in\mathbb{R}^{N_p\times C_v}$ with $N_p=h_v w_v$. During training, we perturb $\mathbf{x}_0$ with a standard forward process. Since the unified 3D token space is high-dimensional, we adopt an $\mathbf{x}_0$-prediction parameterization (as observed effective in JiT~\cite{li2025back}), while computing the training objective in the equivalent $v$-space via a deterministic conversion for numerical stability and compatibility with standard diffusion formulations. The denoiser is conditioned on a single clean conditioning view, its camera parameters, the target camera parameters, and an optional text prompt embedding. Importantly, the conditioning-view tokens are encoded by 3D-URAE independently from the multi-view encoding used to obtain the target token grid $\mathbf{x}_0$, ensuring consistent inference when only the conditioning view is available,

\begin{equation}
\hat{\mathbf{x}}_{0}
=\mathcal{D}_\theta\!\Big(\mathbf{x}_t,\, t,\, \mathbf{c}_0^{(\mathrm{cond})},\, \mathcal{T}^{(\mathrm{cond})},\, \mathcal{T}^{(\mathrm{tgt})},\, \mathbf{e}_{\mathrm{text}}\Big),
\hat{\mathbf{v}}_\theta=\frac{\alpha_t \mathbf{x}_t-\hat{\mathbf{x}}_{0}}{\sigma_t},
\hat{\mathbf{x}}_{0}=\alpha_t \mathbf{x}_t-\sigma_t \hat{\mathbf{v}}_\theta,
\end{equation}
where $\mathbf{c}_0^{(\mathrm{cond})}$ is the flattened clean token grid of the conditioning view and $\mathcal{T}$ denotes the corresponding camera parameters. We optimize the standard $v$-prediction objective:
\begin{equation}
\mathcal{L}_v=\mathbb{E}\!\left[\lVert \hat{\mathbf{v}}_\theta-\mathbf{v}\rVert_2^2\right].
\end{equation}

\textbf{Cross-view Correspondence Preservation Loss.}
While $\mathcal{L}_v$ enforces proximity in token space, it does not explicitly preserve cross-view structural correspondence. We therefore introduce a correspondence-preserving regularizer that aligns the nearest-neighbor matching pattern between the target view and the conditioning view. Concretely, let $\mathbf{c}_0^{(\mathrm{cond})}\in\mathbb{R}^{N_p\times C_v}$ denote the flattened clean tokens of the conditioning view. For each target view token $p$, we compute cosine similarities to all conditioning locations, select the most similar index $q_p^\star$, and keep it only if the confidence exceeds a threshold $\tau=0.9$. We then apply a cross-entropy loss on the correspondence distribution induced by the predicted clean tokens $\hat{\mathbf{x}}_0$,

\begin{equation}
q_p^\star=\arg\max_{q}\ \mathrm{cos}(\mathbf{x}_{0,p},\mathbf{c}^{(\mathrm{cond})}_{0,q}),\qquad
\mathbbm{1}_p=\mathbbm{1}\!\left(\max_{q}\mathrm{cos}(\mathbf{x}_{0,p},\mathbf{c}^{(\mathrm{cond})}_{0,q})\ge \tau\right),
\end{equation}

\begin{equation}
\mathcal{L}_{\mathrm{cvc}}
=\frac{1}{\sum_p \mathbbm{1}_p}\sum_{p=1}^{N_p}
\mathbbm{1}_p\cdot
\Bigg(
-\log
\frac{\exp\!\big(\mathrm{cos}(\hat{\mathbf{x}}_{0,p},\mathbf{c}^{(\mathrm{cond})}_{0,q_p^\star})/T\big)}
{\sum_{q=1}^{N_p}\exp\!\big(\mathrm{cos}(\hat{\mathbf{x}}_{0,p},\mathbf{c}^{(\mathrm{cond})}_{0,q})/T\big)}
\Bigg),
\end{equation}
where $T$ is a temperature hyperparameter. Finally, we combine the two terms:
\begin{equation}
\mathcal{L}_{\mathrm{diff}}=\mathcal{L}_{v}+\lambda_{\mathrm{cvc}}\mathcal{L}_{\mathrm{cvc}}.
\end{equation}

\subsection{Manifold-Drift Forcing}
\label{sec:mdf}

Although the correspondence-preserving diffusion training improves cross-view structural consistency during training, inference still suffers from a train--inference exposure bias: the denoiser is trained to predict from ground-truth noised latents following the forward process, while at inference it conditions on its own intermediate samples. This sampling drift gradually pushes the sampled latents away from the 3D-URAE manifold, and the discrepancy can be amplified in multi-view generation because cross-view constraints couple all views through shared 3D structure (a detailed analysis is provided in the appendix).

To mitigate this issue, we propose a manifold-drift forcing strategy that explicitly trains the downstream 3D decoder to be robust to off-manifold latents produced during sampling. Concretely, for each scene we run the diffusion model to obtain intermediate predictions within a step interval $t\sim\mathcal{U}([T_1,T_2])$ and take the corresponding predicted clean latent $\hat{\mathcal{V}}_{0}^{(t)}$ (\textit{i.e.}, reshaped from $\hat{\mathbf{x}}_0$). Meanwhile, we have the ground-truth unified tokens $\mathcal{V}$ from 3D-URAE. We then construct a drifted training latent by mixing the predicted latent with the ground truth using a ratio $\alpha\in[0,1]$:
\begin{equation}
\tilde{\mathcal{V}}=\alpha\,\hat{\mathcal{V}}_{0}^{(t)}+(1-\alpha)\,\mathcal{V},\qquad
t\sim\mathcal{U}([T_1,T_2]),\ \alpha\sim\mathcal{U}([0,1]).
\end{equation}
We feed $\tilde{\mathcal{V}}$ into the 3D decoder (\textit{i.e.}, the prediction heads) to obtain $(\tilde{\mathcal{G}},\tilde{\mathcal{D}})$ and optimize the same differentiable 3DGS rendering objective as in Sec.~\ref{sec:rae}. Different from the 3D-URAE training, since most views involved in diffusion sampling are already ``noisy'' (including both sampled views and conditioning views), we do not sample additional target views; instead, we render and supervise all available views (the noise views and the conditioning views) under their camera parameters. The objective is
\begin{equation}
\tilde{\mathcal{L}}_{\mathrm{render}}
=\frac{1}{N}\sum_{n=1}^{N}
\Big(
\lVert \tilde{\mathcal{I}}_n-\mathcal{I}^{\mathrm{gt}}_n\rVert_2^2
+\lambda_{\mathrm{lpips}}\ \mathrm{LPIPS}(\tilde{\mathcal{I}}_n,\mathcal{I}^{\mathrm{gt}}_n)
\Big),
\end{equation}
where $\tilde{\mathcal{I}}_n$ is rendered from $(\tilde{\mathcal{G}},\tilde{\mathcal{D}})$ using the $n$-th view camera parameters in the dataset convention. By training the decoder on the interpolated latent $\tilde{\mathcal{V}}$ across a range of sampling steps, the decoder learns to tolerate diffusion-induced manifold drift and yields more stable 3D reconstruction quality at inference.

\begin{figure}[t]
    \centering
    \includegraphics[width=0.9\linewidth]{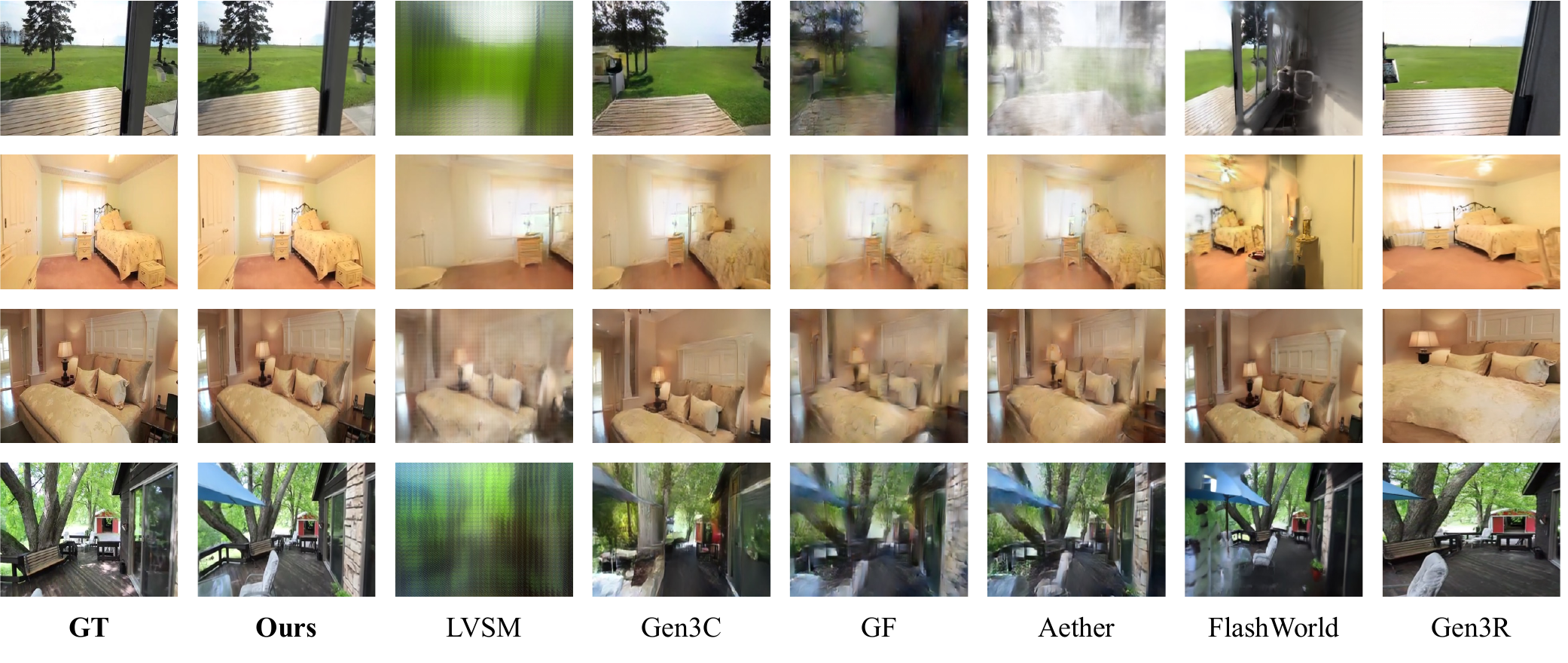}
    \vspace{-3mm}
    \caption{Qualitative visual results on one-view-based novel view generation.}
    \vspace{-3mm}
    \label{fig:vis_compare}
\end{figure}

\section{Experiment}
\subsection{Training Details}
\textbf{Datasets.}
We train on two large-scale calibrated multi-view datasets: RealEstate10K (Re10K)~\cite{zhou2018stereo} and DL3DV-10K~\cite{ling2024dl3dv}. Together they contain $\sim$70K multi-view-consistent scenes covering diverse real-world environments. We use the official train/test split for Re10K. For DL3DV-10K, we randomly split scenes into 90\% for training and 10\% for testing. Scene-level text prompts are generated with a multimodal large language model~\cite{guo2025seed1}.

\textbf{3D-URAE.}
We fine-tune the 3D foundation reconstructor $\pi^3$ to obtain our 3D Unified Representation Autoencoder (3D-URAE). Training uses a 1:1 mixture of Re10K and DL3DV-10K. For each scene, we sample $N{=}8$ views and supervise $N_{\mathrm{novel}}{=}4$ novel views; all images are resized to $224{\times}448$. We optimize a weighted sum of a differentiable 3DGS rendering loss (with $\lambda_{\mathrm{lpips}}{=}0.05$) and a semantic distillation objective (with $\lambda_{\mathrm{sem}}{=}0.1$, $\lambda_{\mathrm{mdms}}{=}1.0$ and margins $m_1{=}m_2{=}0.05$). We initialize from the $\pi^3$ checkpoint and fine-tune with AdamW~\cite{loshchilov2017decoupled}, using a linearly decayed learning rate from $2\times10^{-4}$ to $2\times10^{-5}$. The per-GPU batch size is 2 (global batch size 64). All experiments are run on 32 NVIDIA A100-SXM4-80G GPUs for 30K steps. Additional implementation details and hyperparameter analyses are provided in the Appendix due to space constraints.

\textbf{Diffusion.}
We train a conditional DiT~\cite{peebles2023scalable} denoiser in the unified 3D representation space from 3D-URAE (Sec.~\ref{sec:diffusion_ctcp}). The model predicts $\hat{\mathbf{x}}_0$ and is optimized with the $v$-prediction loss,
$\mathcal{L}_v=\mathbb{E}\!\left[\lVert \hat{\mathbf{v}}_\theta-\mathbf{v}\rVert_2^2\right]$.
We additionally enforce cross-view correspondence preservation on $\hat{\mathbf{x}}_0$,
$\mathcal{L}_{\mathrm{diff}}=\mathcal{L}_v+\lambda_{\mathrm{cvc}}\mathcal{L}_{\mathrm{cvc}}$,
with $\tau{=}0.9$ and $\lambda_{\mathrm{cvc}}{=}0.2$.
We initialize DiT from Wan-2.1-T2I-1.3B and condition on 3D-URAE latents from a single view (with camera parameters). To focus on non-text-conditioned generation, we apply classifier-free text dropping with rate 0.5. Training uses the Re10K and DL3DV-10K mixture with batch size 256 (8 per GPU), linear LR decay $1\times10^{-4}\!\rightarrow\!1\times10^{-5}$, EMA decay 0.9995, and runs for 100K steps. Due to space constraints, additional implementation details and hyperparameter analyses are deferred to the Appendix.

\begin{figure}[t]
  \centering
  \includegraphics[width=0.9\textwidth]{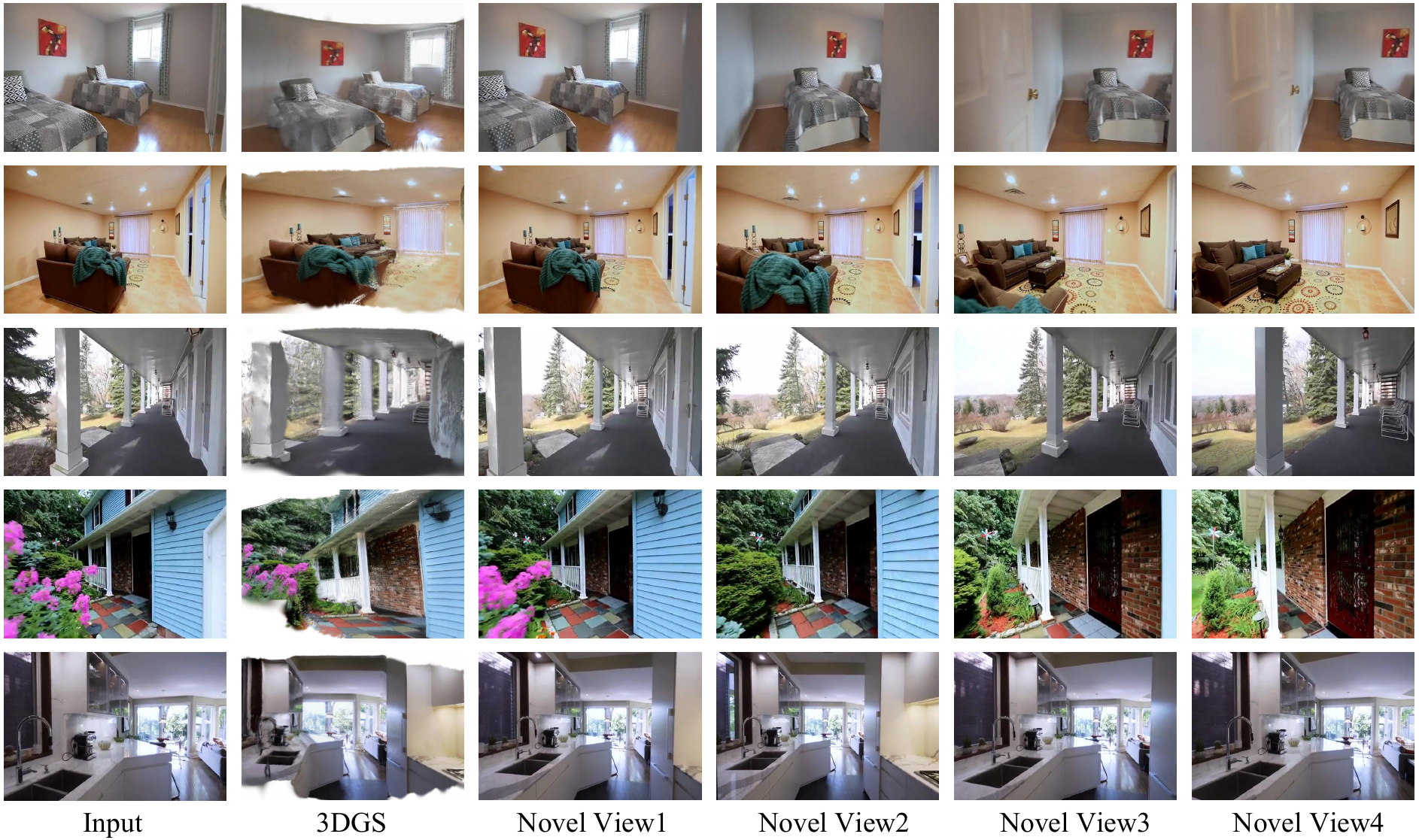}
  \vspace{-2mm}
  \caption{Visualization of 3DGS and rendered novel views.}
  \label{fig:vis_3DGS}
  \vspace{-6mm}
\end{figure}

\textbf{Manifold-drift forcing.}
In the final stage (Sec.~\ref{sec:mdf}), we train the 3D decoder to be robust to off-manifold latents from diffusion sampling. We sample intermediate timesteps $t\sim\mathcal{U}([T_1,T_2])$ with $T_1{=}10$ and $T_2{=}20$, obtain the predicted clean latent $\hat{\mathcal{V}}_{0}^{(t)}$, and form a drifted latent
$\tilde{\mathcal{V}}=\alpha\,\hat{\mathcal{V}}_{0}^{(t)}+(1-\alpha)\,\mathcal{V}$,
where $\alpha\sim\mathcal{U}([0,1])$.
We feed $\tilde{\mathcal{V}}$ into the 3D prediction heads to obtain $(\tilde{\mathcal{G}},\tilde{\mathcal{D}})$ and optimize the same differentiable 3DGS rendering loss as in Sec.~\ref{sec:rae}, supervising all views used for conditioning and sampling. We freeze the 3D-URAE encoder $\mathcal{E}_{\mathcal{V}}$ and update only the decoder heads $\mathcal{D}_{\mathcal{V}}$, generating drifted latents with the DiT from the previous stage. We train with a learning rate of $2\times10^{-5}$, batch size 256 (8 per GPU), for 10K steps.

\subsection{Evaluation Protocols}
\label{sec:eval_protocols}

We evaluate under two protocols: ground-truth-based novel view synthesis (NVS) and the reference-free WorldScore benchmark~\cite{duan2025worldscore}.
In NVS, generation is conditioned only on a single input image together with its camera parameters. We randomly sample 500 scenes from Re10K and 500 scenes from DL3DV-10K, synthesize target views at calibrated camera poses, and compare them with the corresponding ground-truth images. We report PSNR, SSIM~\cite{wang2004image}, and LPIPS~\cite{zhang2018unreasonable} between the generated and ground-truth images, and additionally report the VBench Score~\cite{huang2024vbench,huang2025vbench++} to assess generative capability under image conditioning, focusing on I2V Subject (I2V Subj.), I2V Background (I2V BG), and Imaging Quality (I.Q.).

For WorldScore, each test case provides a single reference image together with a text prompt and camera trajectory, and quality is measured by the standard WorldScore protocol and scoring metrics without paired ground-truth novel views. We follow the benchmark’s photorealistic indoor split (500 scenes) for evaluation. However, the benchmark’s outdoor distribution differs substantially from our training data distribution and our training set is relatively small; therefore, for outdoor evaluation we construct a WorldScore-style set of 500 scenes from DL3DV-10K by treating a single-view image and its text prompt from each outdoor scene as the reference, while matching the camera trajectories from WorldScore to drive generation. We then evaluate using the standard WorldScore protocol and metrics.

We compare against recent baselines spanning view synthesis and geometry-aware world generation: LVSM~\cite{jinlvsm}, a large transformer-based NVS model with minimal explicit 3D inductive bias; Gen3C~\cite{ren2025gen3c}, which improves camera controllability and temporal consistency via an explicit 3D cache; GF (Geometry Forcing)~\cite{wu2025geometry}, which encourages diffusion models to internalize 3D structure for better geometric consistency; Aether~\cite{zhu2025aether}, a geometry-aware unified world modeling framework for joint reconstruction and generation; FlashWorld~\cite{li2025flashworld}, which directly generates 3D Gaussian scene representations for fast rendering-based synthesis; and Gen3R~\cite{huang2026gen3r}, which combines reconstruction and diffusion priors to generate both appearance and geometry.

\begin{wrapfigure}{r}{0.4\textwidth}
  \centering
  \vspace{-5mm}
  \includegraphics[width=0.38\textwidth]{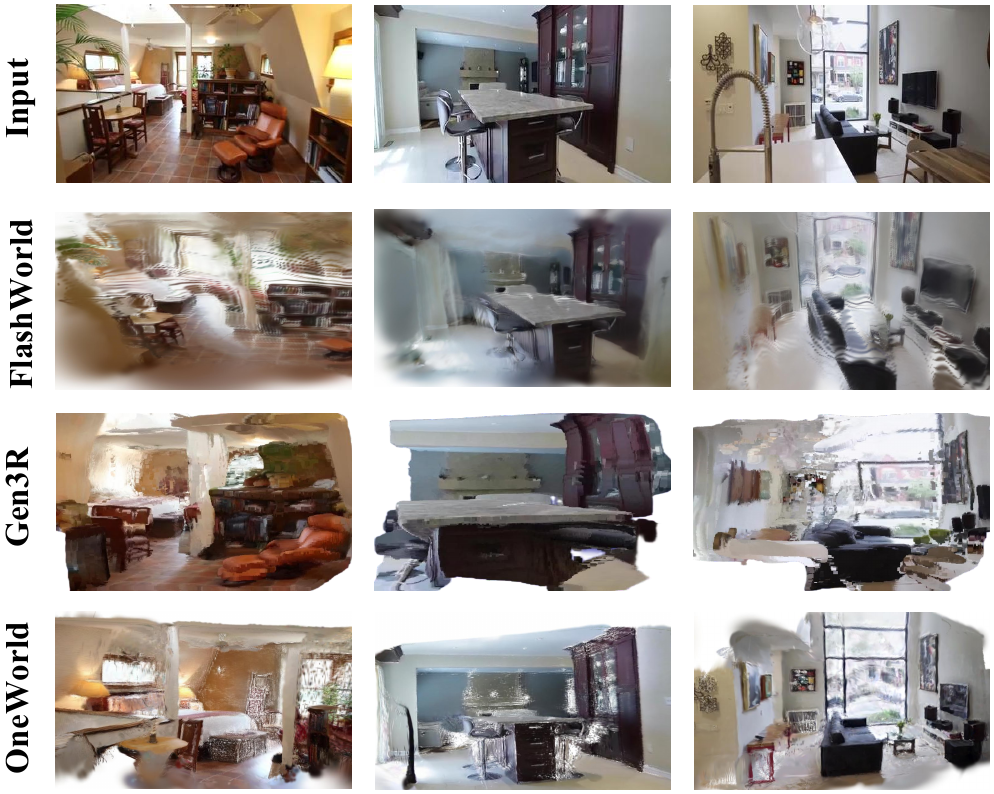}
  \caption{Comparison of 3D scenes generated by different methods: Gen3R~\cite{huang2026gen3r} uses point clouds, while FlashWorld~\cite{li2025flashworld} and OneWorld use 3DGS.}
  \vspace{-5mm}
  \label{fig:vis_3D}
\end{wrapfigure}

\subsection{3D Scene Generation}

\begin{table*}[t]
\centering
\caption{\textbf{Experimental results on 1-view-based novel view generation.}
We report the performance of different methods on the RealEstate10K and DL3DV datasets.}
\vspace{-3mm}
\label{tab:nvs}
\resizebox{\textwidth}{!}{
\setlength{\tabcolsep}{4pt}
\begin{tabular}{l|cccccc|cccccc}
\toprule
\multirow{2}{*}{Method} & \multicolumn{6}{c|}{RealEstate10K} & \multicolumn{6}{c}{DL3DV-10K} \\
\cmidrule(lr){2-7} \cmidrule(lr){8-13}
& PSNR $\uparrow$ & SSIM $\uparrow$ & LPIPS $\downarrow$ & I2V Subj. $\uparrow$ & I2V BG $\uparrow$ & I.Q. $\uparrow$
& PSNR $\uparrow$ & SSIM $\uparrow$ & LPIPS $\downarrow$ & I2V Subj. $\uparrow$ & I2V BG $\uparrow$ & I.Q. $\uparrow$ \\
\midrule
LVSM~\cite{jinlvsm}
& 18.54 & 0.694 & 0.336 & \secondc{0.993} & \thirdc{0.991} & 0.487
& 15.18 & 0.516 & 0.482 & \thirdc{0.962} & 0.966 & 0.421 \\
Gen3C~\cite{ren2025gen3c}
& 19.88 & 0.697 & 0.271 & \thirdc{0.991} & 0.990 & 0.514
& 15.79 & 0.531 & 0.497 & 0.941 & 0.953 & 0.414 \\
GF~\cite{wu2025geometry}
& 15.97 & 0.518 & 0.424 & 0.986 & 0.976 & 0.553
& 11.62 & 0.322 & 0.621 & 0.932 & 0.929 & 0.533 \\
Aether~\cite{zhu2025aether}
& 16.13 & 0.611 & 0.413 & \thirdc{0.991} & 0.989 & 0.536
& 13.37 & 0.492 & 0.566 & 0.957 & 0.963 & 0.451 \\
FlashWorld~\cite{li2025flashworld}
& \secondc{20.18} & \secondc{0.724} & \secondc{0.256} & \secondc{0.993} & \secondc{0.994} & \thirdc{0.584}
& \secondc{16.02} & \secondc{0.566} & \secondc{0.451} & \secondc{0.964} & \thirdc{0.969} & \thirdc{0.538} \\
Gen3R~\cite{huang2026gen3r}
& \thirdc{20.09} & \thirdc{0.714} & \thirdc{0.269} & \bestc{0.994} & \secondc{0.994} & \secondc{0.593}
& \thirdc{15.94} & \thirdc{0.557} & \thirdc{0.468} & \secondc{0.964} & \secondc{0.970} & \secondc{0.543} \\
OneWorld (ours)
& \bestc{21.57} & \bestc{0.735} & \bestc{0.231} & \secondc{0.993} & \bestc{0.995} & \bestc{0.604}
& \bestc{17.19} & \bestc{0.589} & \bestc{0.418} & \bestc{0.966} & \bestc{0.973} & \bestc{0.556} \\
\bottomrule
\end{tabular}
}
\vspace{-4mm}
\end{table*}

\textbf{1-view NVS on calibrated benchmarks.}

We first evaluate 1-view, camera-conditioned novel view generation under the GT-based NVS protocol on RealEstate10K and DL3DV-10K (Tab.~\ref{tab:nvs} for quantitative results; Fig.~\ref{fig:vis_compare} and Fig.~\ref{fig:vis_3DGS} for qualitative visualizations). 
OneWorld achieves the best overall results on both benchmarks. On RealEstate10K, it reaches 21.57 PSNR and 0.735 SSIM with the lowest LPIPS of 0.231, and it also attains the highest image quality score (I.Q. 0.604). Identity consistency is near-saturated, with I2V Subj. 0.993 and I2V BG 0.995. On DL3DV-10K, OneWorld again ranks first, with 17.19 PSNR, 0.589 SSIM, and the lowest LPIPS of 0.418, while improving I2V Subj./BG to 0.966/0.973 and achieving the best I.Q. of 0.556. These results show that OneWorld improves fidelity and perceptual quality while maintaining strong subject and background consistency across diverse real-world scenes. Additionally, we visualize and compare methods capable of generating 3D representations (\textit{e.g.}, point clouds and 3DGS), as shown in Fig.~\ref{fig:vis_3D}. The results indicate that our method surpasses current SOTA baselines in producing coherent 3D scenes.

\begin{table*}[t]
\centering
\caption{WorldScore-style reference-free evaluation results. We report results on WorldScore-Indoor and on DL3DV as an outdoor benchmark.}
\vspace{-3mm}
\label{tab:worldscore}
\small
\renewcommand{\arraystretch}{1.2}
\setlength{\tabcolsep}{6pt}

\resizebox{\textwidth}{!}{
\begin{tabular}{l|cccccc|cccccc}
\toprule
\multirow{2}{*}{\textbf{Method}} &
\multicolumn{6}{c|}{WorldScore-Indoor} &
\multicolumn{6}{c}{DL3DV} \\
\cmidrule(lr){2-7} \cmidrule(lr){8-13}
& \makecell{3D\\ Consist.} &
  \makecell{Photo.\\ Consist.} &
  \makecell{Obj.\\ Cont.} &
  \makecell{Cont.\\ Align.} &
  \makecell{Style\\ Consist.} &
  \makecell{Subj.\\ Quality} &
  \makecell{3D\\ Consist.} &
  \makecell{Photo.\\ Consist.} &
  \makecell{Obj.\\ Cont.} &
  \makecell{Cont.\\ Align.} &
  \makecell{Style\\ Consist.} &
  \makecell{Subj.\\ Quality} \\
\midrule

LVSM~\cite{jinlvsm}
& 74.32 & 63.94 & 47.21 & 23.79 & 66.02 & 36.38
& 64.37 & 55.58 & 40.21 & 18.88 & 57.65 & 29.59 \\
Gen3C~\cite{ren2025gen3c}
& 74.11 & 78.74 & \bestc{51.36} & 27.12 & 71.18 & 40.05
& 69.41 & 69.96 & \bestc{44.18} & 21.92 & 62.23 & 33.07 \\
GF~\cite{wu2025geometry}
& 78.05 & 69.21 & 46.95 & 24.62 & 68.71 & 38.66
& 67.12 & 58.44 & 41.67 & 20.35 & 59.92 & 31.74 \\
Aether~\cite{zhu2025aether}
& 77.82 & 66.66 & \thirdc{49.28} & 25.04 & 69.06 & 38.17
& 67.47 & 57.98 & 42.02 & 19.83 & 60.37 & 31.11 \\
FlashWorld~\cite{li2025flashworld}
& \secondc{83.57} & \secondc{80.19} & \secondc{49.61} & \bestc{49.27} & \secondc{75.32} & \bestc{54.09}
& \secondc{76.74} & \secondc{72.76} & \secondc{43.25} & \bestc{42.32} & \secondc{69.08} & \bestc{46.62} \\
Gen3R~\cite{huang2026gen3r}
& \thirdc{82.12} & \thirdc{78.83} & 48.06 & \secondc{47.71} & \thirdc{73.89} & \secondc{52.58}
& \thirdc{75.29} & \thirdc{71.44} & 41.86 & \secondc{40.95} & \thirdc{67.63} & \thirdc{45.09} \\
OneWorld (ours)
& \bestc{84.98} & \bestc{81.67} & 48.92 & \thirdc{46.88} & \bestc{76.74} & \thirdc{51.73}
& \bestc{78.21} & \bestc{74.09} & \thirdc{42.58} & \thirdc{40.12} & \bestc{70.62} & \secondc{45.98} \\
\bottomrule
\end{tabular}
}
\end{table*}

\textbf{WorldScore-style reference-free evaluation.}
We further assess single-image world generation with the reference-free WorldScore protocol (Tab.~\ref{tab:worldscore}). On WorldScore-Indoor, OneWorld achieves the best 3D Consistency at 84.98, Photometric Consistency at 81.67, and Style Consistency at 76.74, indicating stronger multi-view coherence and more stable appearance along the trajectory. For Object Control, Content Alignment, and Subjective Quality, OneWorld remains competitive with scores of 48.92, 46.88, and 51.73, although other methods achieve the top score on these axes in some cases. On the DL3DV outdoor benchmark, OneWorld again ranks first in 3D Consistency, reaching 78.21, along with the best Photometric Consistency of 74.09 and Style Consistency of 70.62. It also delivers strong performance on Object Control, Content Alignment, and Subjective Quality with scores of 42.58, 40.12, and 45.98, staying close to the best baselines. Overall, the reference-free results show that OneWorld improves long-range 3D and appearance consistency across both indoor and outdoor settings, while maintaining competitive controllability and perceptual quality.

\subsection{Ablation Study}
\begin{figure}[t]
  \centering
  \includegraphics[width=0.9\textwidth]{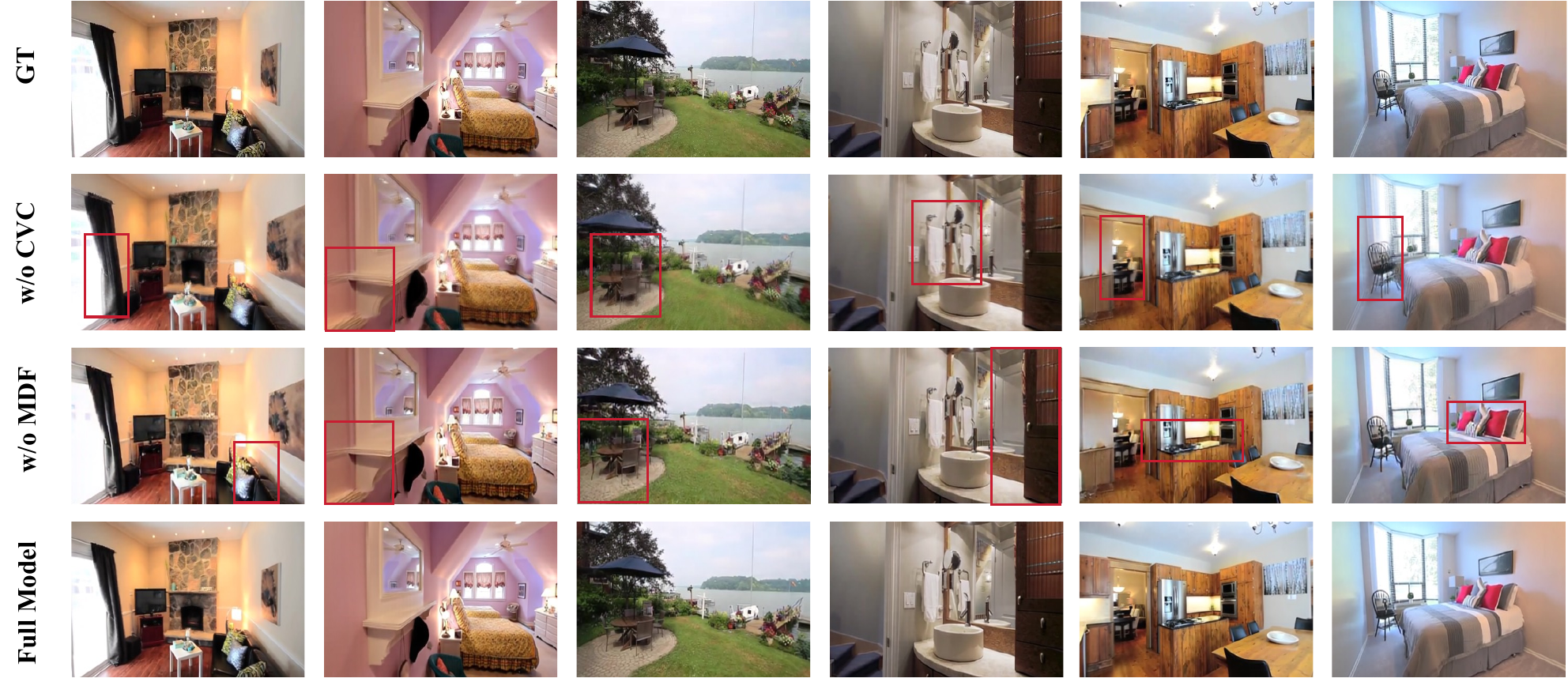}
  \caption{\textbf{Qualitative visualization of the ablation study under the 1-view-based NVS setting.} We compare the full model with variants without Cross-View Correspondence and without Manifold-Drift Forcing.}
  \label{fig:ablation_gen}
\end{figure}

As shown in Tab.~\ref{tab:ablation_re10k_1view} and Fig.~\ref{fig:ablation_gen}, both components contribute substantially to view-consistent generation quality. Removing the correspondence regularizer (w/o CVC) leads to a clear degradation across all appearance metrics, with PSNR dropping from 21.57 to 19.10, SSIM decreasing from 0.735 to 0.682, and LPIPS increasing from 0.231 to 0.284. This verifies that explicitly enforcing cross-view correspondence during diffusion training is crucial for preserving multi-view alignment in the unified 3D representation space, and helps suppress view-dependent artifacts that otherwise accumulate when synthesizing novel viewpoints from a single observation.

We also observe a consistent, though milder, degradation when removing manifold-drift forcing (w/o MDF). Without MDF, the reconstruction scores drop from 21.57 to 20.59 and from 0.735 to 0.714, while the perceptual distance increases from 0.231 to 0.256. This suggests that the 3D decoding heads become more sensitive to the shifted latents encountered along the sampling trajectory. Overall, the full model achieves the best fidelity and perceptual quality. The results imply that CVC mainly improves cross-view structural consistency during generation, whereas MDF further stabilizes decoding under the inference-time shift, leading to more reliable renderings in the 1-view NVS setting.

\begin{table}[t]
\centering
\caption{Ablation on generation components evaluated on RealEstate10K under the 1-view NVS setting.}
\label{tab:ablation_re10k_1view}
\setlength{\tabcolsep}{8pt}
\resizebox{0.8\linewidth}{!}{
\begin{tabular}{l|cccccc}
\toprule
Method & PSNR $\uparrow$ & SSIM $\uparrow$ & LPIPS $\downarrow$ & I2V Subj. $\uparrow$ & I2V BG $\uparrow$ & I.Q. $\uparrow$ \\
\midrule
OneWorld (full) & 21.57 & 0.735 & 0.231 & 0.993 & 0.995 & 0.604 \\
w/o CVC & 19.10 & 0.682 & 0.284 & 0.989 & 0.990 & 0.566 \\
w/o MDF & 20.59 & 0.714 & 0.256 & 0.992 & 0.994 & 0.589 \\
\bottomrule
\end{tabular}
}
\end{table}

\section{Conclusion}

We present OneWorld, a diffusion framework that generates 3D scenes directly in the representation space of a pretrained 3D foundation model, overcoming the structural inconsistency and inefficiency of per-view 2D latent pipelines.
By unifying geometry, appearance, and semantics in a coherent 3D representation space and explicitly enforcing cross-view structure while stabilizing sampling, OneWorld enables consistent and high-fidelity 3D scene generation.
Our findings in this paper suggest that generative modeling in 3D foundation representation space marks a promising paradigm toward scalable and unified 3D world generation.

\clearpage
\appendix
In the appendix, we provide complementary details and analyzes that are omitted from the main paper due to space limits.
Specifically, we include:

\begin{itemize}
    \item \textbf{Training hyperparameter settings (Appendix~\ref{sec:hyperparams}).}
    We analyze the impact of key hyperparameters in 3D-URAE reconstruction and unified-space diffusion on performance.

    \item \textbf{Prediction space comparison (Appendix~\ref{sec:pred_space}).}
    We compare $\mathbf{x}_0$-prediction and $v$-prediction in the high-dimensional 3D-URAE token space and analyze their convergence behavior.

    \item \textbf{Feed-forward 3D Gaussian Splatting reconstruction (Appendix~\ref{sec:ffgs}).}
    We compare our method with current state-of-the-art approaches under the same 8-view setting.

    \item \textbf{Proof sketch for manifold-drift forcing (Appendix~\ref{sec:proof}).}
    We give a concise theoretical sketch showing why train--inference mismatch induces off-manifold drift during sampling, why the effect can be amplified in multi-view generation, and how manifold-drift forcing improves decoder robustness.

    \item \textbf{More visualizations (Appendix~\ref{sec:more_vis}).}
    We present additional qualitative results to better illustrate generation behaviors across different scenes and viewpoints.

    \item \textbf{Discussion (Appendix~\ref{sec:discussion}).}
    We summarize limitations and clarify LLM usage details for transparency.
\end{itemize}

\section{Training hyperparameter settings}
\label{sec:hyperparams}

We report the training hyperparameter settings for 3D-URAE reconstruction and unified-space diffusion (Tab.~\ref{tab:ablation_urae_sem_hyper} and Tab.~\ref{tab:ablation_re10k_1view_cvc_hyper}).
Unless otherwise specified, the core training configuration is kept constant (optimizer, learning rate, batch size, token sizes, and rendering-loss weights), while we explore a set of ablation-specific hyperparameters.
We evaluate the resulting changes in 3D-URAE reconstruction performance and OneWorld generation performance using PSNR, SSIM, LPIPS, and semantic similarity.

\subsection{3D-URAE reconstruction hyperparameters}
\label{sec:hyperparams_urae}

3D-URAE is trained with a rendering loss and a semantic distillation loss,
$\mathcal{L}_{\mathrm{URAE}}=\mathcal{L}_{\mathrm{render}}+\lambda_{\mathrm{sem}}\mathcal{L}_{\mathrm{sem}}$. 
In this section, we study the two semantic distillation knobs varied in Tab.~\ref{tab:ablation_urae_sem_hyper}: the margins $m_1$ and $m_2$ (we set $m_1=m_2=m$) and the semantic weight $\lambda_{\mathrm{sem}}$. 
We use $m=0.05$ and $\lambda_{\mathrm{sem}}=0.10$ as default because it gives strong reconstruction and high semantic similarity, where the semantic similarity is measured as the feature similarity to the distilled DINOv2~\cite{oquab2024dinov2} targets.

Tab.~\ref{tab:ablation_urae_sem_hyper} reveals consistent trends. 
With $\lambda_{\mathrm{sem}}=0.10$, increasing the margin from $m=0.00$ to $m=0.05$ boosts PSNR from 26.67 to 28.19 and SSIM from 0.918 to 0.932, while reducing LPIPS from 0.122 to 0.102. 
Meanwhile, semantic similarity remains high, changing only slightly from 0.986 to 0.984. 
However, when the margin is further increased to $m=0.10$, semantic similarity drops sharply from 0.984 to 0.917, and reconstruction quality also degrades mildly, with SSIM decreasing from 0.932 to 0.930 and LPIPS increasing from 0.102 to 0.105. 
These results suggest that a small margin helps the semantic loss emphasize meaningful mismatches, whereas an overly large margin imposes an excessively strict constraint that hinders learning. We observe a similar trade-off when varying the semantic weight while fixing $m=0.05$. 
Setting $\lambda_{\mathrm{sem}}=0.05$ yields the best reconstruction scores, reaching PSNR 28.31, SSIM 0.933, and LPIPS 0.101, but it comes with a lower semantic similarity of 0.947. 
Increasing the weight to $\lambda_{\mathrm{sem}}=0.10$ keeps reconstruction nearly unchanged, with PSNR 28.19, SSIM 0.932, and LPIPS 0.102, while substantially improving semantic similarity to 0.984. 
Further increasing to $\lambda_{\mathrm{sem}}=0.20$ leads to clear reconstruction deterioration, where PSNR falls to 26.58, SSIM to 0.916, and LPIPS rises to 0.125, despite maintaining a high semantic similarity of 0.987, indicating over-regularization.

\begin{table*}[t]
\centering
\caption{\textbf{Detailed ablations on semantic distillation hyperparameters.}
Left: margins with $m_1{=}m_2$ (fix $\lambda_{\mathrm{sem}}{=}0.1$). Right: semantic weight $\lambda_{\mathrm{sem}}$ (fix $m_1{=}m_2{=}0.05$). We report PSNR/SSIM/LPIPS and semantic similarity (Sem. Sim.).}
\label{tab:ablation_urae_sem_hyper}
\resizebox{0.95\textwidth}{!}{
\setlength{\tabcolsep}{4pt}
\begin{tabular}{c|cccc|c|cccc}
\toprule
\multicolumn{5}{c|}{Margins ($m_1{=}m_2{=}m$, fix $\lambda_{\mathrm{sem}}{=}0.1$)} &
\multicolumn{5}{c}{Semantic weight (fix $m_1{=}m_2{=}0.05$)} \\
\cmidrule(lr){1-5} \cmidrule(lr){6-10}
$m$ & PSNR $\uparrow$ & SSIM $\uparrow$ & LPIPS $\downarrow$ & Sem. Sim. $\uparrow$
& $\lambda_{\mathrm{sem}}$ & PSNR $\uparrow$ & SSIM $\uparrow$ & LPIPS $\downarrow$ & Sem. Sim. $\uparrow$ \\
\midrule
0.00 & 26.67 & 0.918 & 0.122 & 0.986
& 0.05 & 28.31 & 0.933 & 0.101 & 0.947 \\
0.05 & 28.19 & 0.932 & 0.102 & 0.984
& 0.10 & 28.19 & 0.932 & 0.102 & 0.984 \\
0.10 & 28.19 & 0.930 & 0.105 & 0.917
& 0.20 & 26.58 & 0.916 & 0.125 & 0.987 \\
\bottomrule
\end{tabular}
}
\end{table*}

\subsection{Diffusion and generation hyperparameters}
\label{sec:hyperparams_diffusion}

\begin{table}[t]
\centering
\caption{\textbf{Ablation on CVC hyperparameters evaluated on RealEstate10K under the 1-view NVS setting.}
Top: threshold $\tau$ (fix $\lambda_{\mathrm{cvc}}{=}0.2$). Bottom: CVC weight $\lambda_{\mathrm{cvc}}$ (fix $\tau{=}0.9$).}
\vspace{2mm}
\label{tab:ablation_re10k_1view_cvc_hyper}
\setlength{\tabcolsep}{8pt}
\resizebox{0.95\linewidth}{!}{
\begin{tabular}{l|cccccc}
\toprule
Setting & PSNR $\uparrow$ & SSIM $\uparrow$ & LPIPS $\downarrow$ & I2V Subj. $\uparrow$ & I2V BG $\uparrow$ & I.Q. $\uparrow$ \\
\midrule
\multicolumn{7}{c}{Ablation on threshold $\tau$ \ \ (fix $\lambda_{\mathrm{cvc}}{=}0.2$)} \\
\midrule
$\tau{=}0.80$ & 17.62 & 0.645 & 0.316 & 0.976 & 0.981 & 0.541 \\
$\tau{=}0.90$ & 19.10 & 0.682 & 0.284 & 0.989 & 0.990 & 0.566 \\
$\tau{=}0.95$ & 18.21 & 0.661 & 0.301 & 0.982 & 0.986 & 0.553 \\
\midrule
\multicolumn{7}{c}{Ablation on CVC weight $\lambda_{\mathrm{cvc}}$ \ \ (fix $\tau{=}0.9$)} \\
\midrule
$\lambda_{\mathrm{cvc}}{=}0.10$ & 18.05 & 0.657 & 0.305 & 0.981 & 0.986 & 0.552 \\
$\lambda_{\mathrm{cvc}}{=}0.20$ & 19.10 & 0.682 & 0.284 & 0.989 & 0.990 & 0.566 \\
$\lambda_{\mathrm{cvc}}{=}0.40$ & 17.41 & 0.639 & 0.323 & 0.973 & 0.979 & 0.538 \\
\bottomrule
\end{tabular}
}
\vspace{-3mm}
\end{table}

We train a conditional diffusion model in the unified token space using a standard velocity loss $\mathcal{L}_v$ and a cross-view correspondence (CVC) loss,
$\mathcal{L}_{\mathrm{diff}}=\mathcal{L}_v+\lambda_{\mathrm{cvc}}\mathcal{L}_{\mathrm{cvc}}$.
In this section, we analyze the two CVC hyperparameters varied in Tab.~\ref{tab:ablation_re10k_1view_cvc_hyper}: the threshold $\tau$ for keeping only confident matches and the loss weight $\lambda_{\mathrm{cvc}}$.
To reduce the training cost of hyperparameter exploration, all experiments in this table are trained for 10K steps.
We use $\tau=0.90$ and $\lambda_{\mathrm{cvc}}=0.20$ as default.

Tab.~\ref{tab:ablation_re10k_1view_cvc_hyper} indicates that both $\tau$ and $\lambda_{\mathrm{cvc}}$ admit an effective operating range. 
When fixing $\lambda_{\mathrm{cvc}}=0.20$, setting the threshold to $\tau=0.80$ yields PSNR 17.62, SSIM 0.645, and LPIPS 0.316. 
Raising the threshold to $\tau=0.90$ improves all three metrics to PSNR 19.10, SSIM 0.682, and LPIPS 0.284, and also achieves the best perceptual scores, with I2V Subj. 0.989, I2V BG 0.990, and I.Q. 0.566. 
Further increasing the threshold to $\tau=0.95$ reduces performance, where PSNR decreases to 18.21, SSIM to 0.661, and LPIPS increases to 0.301. 
This behavior is consistent with the role of thresholding: a lower threshold admits more incorrect matches, whereas an overly high threshold leaves too few correspondences to provide reliable structural guidance. A similar pattern holds when varying the correspondence weight while fixing $\tau=0.90$. 
Using $\lambda_{\mathrm{cvc}}=0.10$ is insufficient, producing PSNR 18.05, SSIM 0.657, and LPIPS 0.305. 
Increasing to $\lambda_{\mathrm{cvc}}=0.20$ gives the best overall results, reaching PSNR 19.10, SSIM 0.682, and LPIPS 0.284. 
Further increasing the weight to $\lambda_{\mathrm{cvc}}=0.40$ degrades quality, with PSNR dropping to 17.41, SSIM to 0.639, and LPIPS rising to 0.323. 
One plausible explanation is that an overly large weight can over-enforce correspondences in challenging regions such as occlusions or large viewpoint changes, which ultimately reduces fidelity.

\section{Prediction Space Comparison}
\label{sec:pred_space}

\begin{figure}[t]
    \centering
    \includegraphics[width=0.95\linewidth]{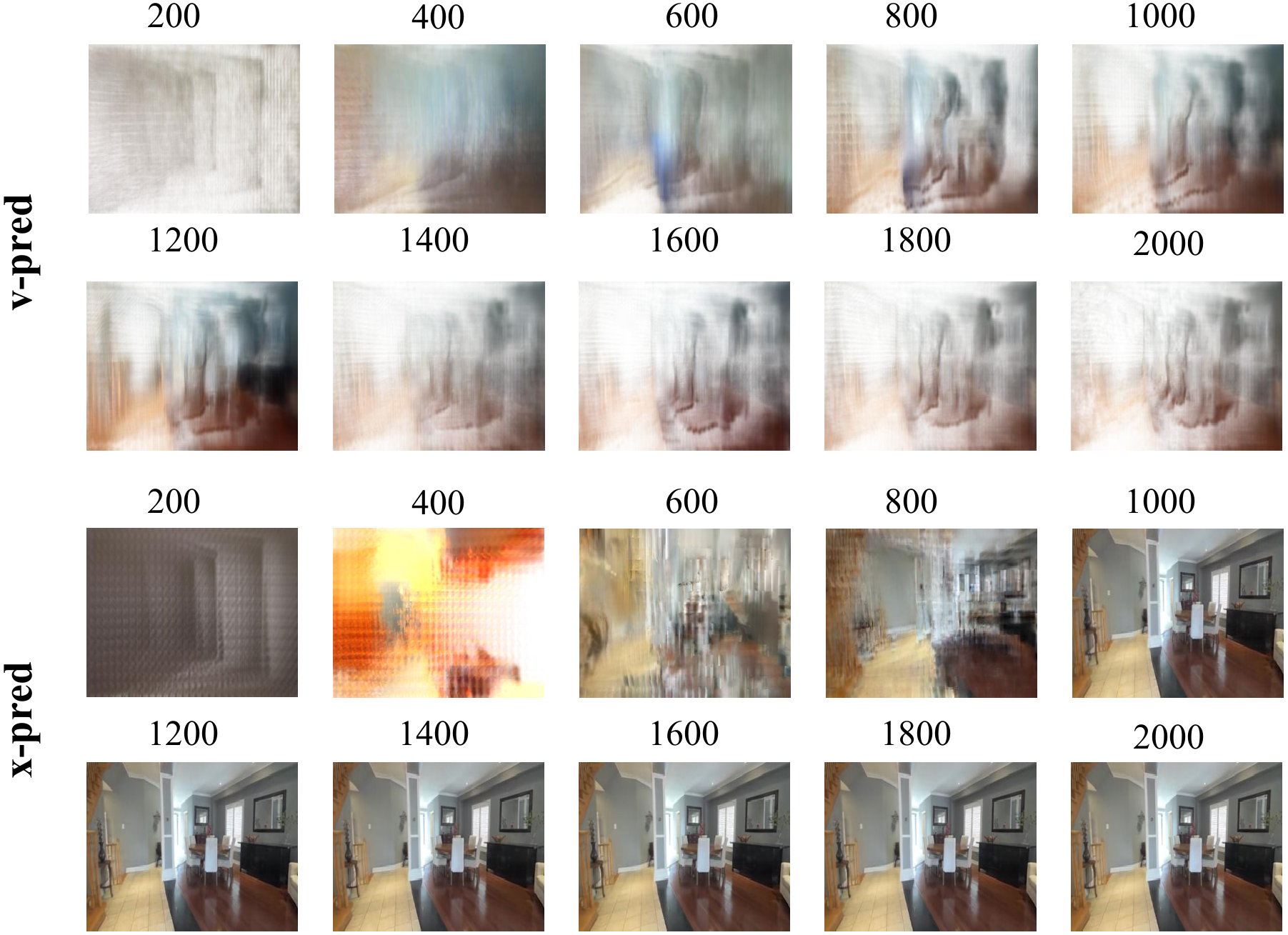}
    \vspace{-1mm}
    \caption{\textbf{Prediction space comparison on single-scene overfitting.}
    We overfit a single training scene for 2000 iterations and visualize the fitting results at different training steps.
    Compared to $v$-prediction, $\mathbf{x}_0$-prediction converges faster and reaches higher-fidelity structure and appearance under the same optimization budget.}
    \label{fig:pred_space}
    \vspace{-4mm}
\end{figure}

Our diffusion model operates on 3D-URAE tokens, whose ambient dimension is high (each token is a high-dimensional feature vector rather than a heavily-compressed VAE latent).
In such a high-dimensional feature space, the choice of prediction target becomes critical.
Following JiT~\cite{li2025back}, we adopt the manifold assumption: although the data are represented in a high-dimensional ambient space, clean data $\mathbf{x}_0$ concentrates near a low-dimensional manifold that captures the underlying structure.
In our case, 3D-URAE tokens are explicitly regularized to align with DINOv2~\cite{oquab2024dinov2} semantics and 3D geometry, which further enforces structured correlations and makes the clean-token distribution more ``manifold-like''.

In contrast, the diffusion noise $\boldsymbol{\epsilon}$ (and the velocity target $v$, which is a linear combination involving $\boldsymbol{\epsilon}$) does not lie on this manifold and instead spreads across the full ambient space.
As argued in JiT~\cite{li2025back}, predicting such high-dimensional noised quantities can be substantially harder than predicting clean data, and may even fail catastrophically as the ambient dimension grows.
Recent theory that revisits prediction targets through dimensionality provides a complementary explanation: when the ambient dimension $D$ significantly exceeds the intrinsic dimension $d$ of the data manifold, the optimal prediction target shifts toward $\mathbf{x}_0$ rather than $v$.
Consistent with these insights, Fig.~\ref{fig:pred_space} shows that $\mathbf{x}_0$-prediction fits a single scene faster and with fewer residual artifacts than $v$-prediction under identical training steps.
\section{Feed-forward 3D Gaussian Splatting Reconstruction}
\label{sec:ffgs}
We evaluate feed-forward 3D Gaussian Splatting reconstruction on RealEstate10K~\cite{zhou2018stereo} and DL3DV~\cite{ling2024dl3dv}. We compare against recent representative feed-forward 3DGS methods, including PixelSplat~\cite{charatan2024pixelsplat} and MVSplat~\cite{chen2024mvsplat} as strong multi-view splatting baselines, DepthSplat~\cite{xu2025depthsplat} as a depth-aware design that achieves the strongest numbers among prior methods in our table, and AnySplat~\cite{jiang2025anysplat} as a general feed-forward splatting baseline. Following our training and evaluation setting, each method reconstructs a 3DGS scene from 8 input views with dataset-provided camera parameters, then renders novel views for evaluation. We report PSNR, SSIM, and LPIPS computed on the rendered novel-view images, where higher PSNR and SSIM and lower LPIPS indicate better quality.

Tab.~\ref{tab:ffgs_compare} shows that 3D-URAE consistently improves novel-view rendering quality under the same 8-view reconstruction setting. 
On RE10K~\cite{zhou2018stereo}, the previous best baseline is DepthSplat~\cite{xu2025depthsplat} with PSNR 27.02 and SSIM 0.879, while 3D-URAE reaches PSNR 28.19 and SSIM 0.932. 
For perceptual similarity, the best baseline LPIPS lies between 0.169 and 0.177, whereas 3D-URAE reduces LPIPS to 0.102. 
On DL3DV~\cite{ling2024dl3dv}, 3D-URAE obtains the best PSNR of 24.68 and SSIM of 0.816, and achieves LPIPS 0.140, which is close to the best value 0.135 from DepthSplat~\cite{xu2025depthsplat}. 
These improvements support the core claim of our 3D Unified Representation Autoencoder: by injecting appearance cues and distilling semantic structure into geometry-aware tokens, the resulting unified 3D latents are more renderable and perceptually faithful, making them a stronger geometry-aware representation basis for subsequent diffusion modeling.

\begin{table*}[t]
\centering
\caption{\textbf{Feed-forward 3D Gaussian Splatting comparison.}
We report PSNR, SSIM, and LPIPS on RealEstate10K and DL3DV.}
\vspace{-3mm}
\label{tab:ffgs_compare}
\resizebox{0.8\textwidth}{!}{
\setlength{\tabcolsep}{6pt}
\begin{tabular}{l|ccc|ccc}
\toprule
\multirow{2}{*}{Method} &
\multicolumn{3}{c|}{RealEstate10K} &
\multicolumn{3}{c}{DL3DV} \\
\cmidrule(lr){2-4} \cmidrule(lr){5-7}
& PSNR $\uparrow$ & SSIM $\uparrow$ & LPIPS $\downarrow$
& PSNR $\uparrow$ & SSIM $\uparrow$ & LPIPS $\downarrow$ \\
\midrule
PixelSplat~\cite{charatan2024pixelsplat}
& \thirdc{26.08} & 0.871 & \secondc{0.169}
& 22.26 & 0.753 & 0.179 \\
MVSplat~\cite{chen2024mvsplat}
& 26.02 & 0.872 & 0.177
& \thirdc{22.61} & 0.759 & \secondc{0.174} \\
DepthSplat~\cite{xu2025depthsplat}
& \secondc{27.02} & \secondc{0.879} & \thirdc{0.174}
& \secondc{23.39} & \secondc{0.796} & \bestc{0.135} \\
AnySplat~\cite{jiang2025anysplat}
& 25.97 & \thirdc{0.873} & 0.181
& 21.12 & 0.736 & 0.204 \\
3D-URAE
& \bestc{28.19} & \bestc{0.932} & \bestc{0.102}
& \bestc{24.68} & \bestc{0.816} & \thirdc{0.140} \\
\bottomrule
\end{tabular}
}
\end{table*}
\section{Proof Sketch for Manifold-Drift Forcing}
\label{sec:proof}

\noindent\textbf{Setup.}
For a scene, the 3D-URAE outputs unified tokens $\mathcal{V}\in\mathbb{R}^{N\times h_v\times w_v\times C_v}$. To model multi-view generation consistently, we sample $N$ target-view token grids and flatten them into a joint latent state:
\begin{equation}
\mathbf{X}_0 = [\mathbf{x}_0^{(1)}; \dots; \mathbf{x}_0^{(N)}] \in \mathbb{R}^{N \times N_p \times C_v},
\end{equation}
where $N_p = h_v w_v$. Let $\mathcal{M}\subset\mathbb{R}^{N \times N_p \times C_v}$ denote the set of valid, 3D-consistent multi-view latents induced by 3D-URAE~\cite{shi2023mvdream}.
The diffusion denoiser is $\hat{\mathbf{X}}_{0} = \mathcal{D}_\theta(\mathbf{X}_t, t, \mathbf{y})$, where the conditioning tuple is $\mathbf{y} := (\mathbf{c}_0^{(\mathrm{cond})}, \mathcal{T}^{(\mathrm{cond})}, \mathcal{T}^{(\mathrm{tgt})}, \mathbf{e}_{\mathrm{text}})$.
Let $S(\cdot)$ denote a sampler update that maps $(\mathbf{X}_t, \hat{\mathbf{X}}_0, t)$ to $\mathbf{X}_{t-1}$. Define the induced one-step sampling map:
\begin{equation}
F_\theta(\mathbf{X}_t, t, \mathbf{y}) := S\big(\mathbf{X}_t,\, \mathcal{D}_\theta(\mathbf{X}_t, t, \mathbf{y}),\, t\big).
\end{equation}
Inference performs the rollout $\tilde{\mathbf{X}}_{t-1} = F_\theta(\tilde{\mathbf{X}}_t, t, \mathbf{y})$ for $t = T_{\mathrm{diff}}, \dots, 1$.

\noindent\textbf{Claim 1.}
The train and inference input distributions for $\mathcal{D}_\theta$ are different.
Training evaluates inputs drawn from the forward noising of $\mathbf{X}_0 \in \mathcal{M}$.
Inference evaluates $\mathcal{D}_\theta$ on $\tilde{\mathbf{X}}_t$ produced by the model rollout.
This difference yields accumulated rollout error and increases the distance of $\tilde{\mathbf{X}}_0$ to $\mathcal{M}$~\cite{ho2020denoising,songscore,karras2022elucidating,songdenoising}.

Let $F^\star(\cdot, t, \mathbf{y})$ be an oracle one-step map such that the oracle chain satisfies $\mathbf{X}_{t-1}^\star = F^\star(\mathbf{X}_t^\star, t, \mathbf{y})$ with $\mathbf{X}_0^\star \in \mathcal{M}$.
Assume the Lipschitz condition $\|F_\theta(\mathbf{A}, t, \mathbf{y}) - F_\theta(\mathbf{B}, t, \mathbf{y})\| \le L_t \|\mathbf{A} - \mathbf{B}\|$ for all joint latents $\mathbf{A}, \mathbf{B}$, and the one-step error bound on oracle inputs $\|F_\theta(\mathbf{X}_t^\star, t, \mathbf{y}) - F^\star(\mathbf{X}_t^\star, t, \mathbf{y})\| \le \varepsilon_t$. Let $\delta_t := \|\tilde{\mathbf{X}}_t - \mathbf{X}_t^\star\|$.

\noindent\textbf{Proof.}
We start from $\|\tilde{\mathbf{X}}_{t-1} - \mathbf{X}_{t-1}^\star\| = \|F_\theta(\tilde{\mathbf{X}}_t, t, \mathbf{y}) - F^\star(\mathbf{X}_t^\star, t, \mathbf{y})\|$. We add and subtract $F_\theta(\mathbf{X}_t^\star, t, \mathbf{y})$ and apply the triangle inequality to obtain the recursion:
\begin{equation}
\delta_{t-1} \le L_t\,\delta_t + \varepsilon_t.
\end{equation}
Unrolling yields a bound on $\delta_0$. Define the manifold distance $d(\tilde{\mathbf{X}}_0, \mathcal{M}) := \inf_{\mathbf{M} \in \mathcal{M}} \|\tilde{\mathbf{X}}_0 - \mathbf{M}\|$. Since $\mathbf{X}_0^\star \in \mathcal{M}$, we have $d(\tilde{\mathbf{X}}_0, \mathcal{M}) \le \delta_0$, yielding:
\begin{equation}
d(\tilde{\mathbf{X}}_0, \mathcal{M}) \le \Big(\prod_{k=1}^{T_{\mathrm{diff}}} L_k\Big)\delta_{T_{\mathrm{diff}}} + \sum_{t=1}^{T_{\mathrm{diff}}}\Big(\prod_{k=1}^{t-1} L_k\Big)\varepsilon_t.
\end{equation}

\noindent\textbf{Claim 2.}
Drift is amplified in unified 3D multi-view generation.
A perturbation in the latent can propagate across views through coupled denoising dynamics~\cite{liu2023syncdreamer,shi2023mvdream,liu2023zero}.
The same perturbation can degrade multiple rendered views through the shared 3D decoder.

Write the multi-view step as $\mathbf{X}_{t-1} = F_\theta(\mathbf{X}_t, t, \mathbf{y})$. Let $J_t = \partial F_\theta / \partial \mathbf{X}_t$ be block-structured. Assume $\|J_{ii}\| \le \kappa_t$ and $\|J_{ij}\| \le \rho_t$ for $i \neq j$. The Lipschitz constant $L_t$ is bounded by the matrix norm:
\begin{equation}
L_t \approx \|J_t\| \le \max_i \sum_{j=1}^N \|J_{ij}\| \le \kappa_t + (N-1)\rho_t.
\end{equation}

\noindent\textbf{Proof.} 
Substituting Eq.~(5) into Eq.~(4) reveals that the accumulated error $\delta_0$ grows with the number of coupled views $N$. In independent generation ($\rho_t = 0$), $L_t \le \kappa_t$. In coupled 3D generation ($\rho_t > 0$), cross-view attention explicitly inflates $L_t$, amplifying the drift.

Let $\mathcal{D}_{\mathcal{V}}$ map the joint latent $\mathbf{X}$ to $(\mathcal{G},\mathcal{D})$. Let $\mathcal{R}(\cdot; \mathcal{T}^{(n)})$ denote differentiable rendering for view $n$, and define $f_n(\mathbf{X}) := \mathcal{R}\big(\mathcal{D}_{\mathcal{V}}(\mathbf{X});\ \mathcal{T}^{(n)}\big)$. Assume each $f_n$ is $K_n$-Lipschitz: $\|f_n(\mathbf{A}) - f_n(\mathbf{B})\| \le K_n\|\mathbf{A} - \mathbf{B}\|$. For any shared off-manifold drift $\Delta\mathbf{X}$, the total visual discrepancy is:
\begin{equation}
\sum_{n=1}^{N} \|f_n(\mathbf{X}+\Delta\mathbf{X}) - f_n(\mathbf{X})\| \le \Big(\sum_{n=1}^{N} K_n\Big)\,\|\Delta\mathbf{X}\|.
\end{equation}
Because $\mathcal{D}_{\mathcal{V}}$ is entirely shared, a local artifact $\Delta\mathbf{X}$ is universally projected into all $N$ views.

\noindent\textbf{Implication for manifold-drift forcing.}
The 3D decoder $\mathcal{D}_{\mathcal{V}}$ is trained mainly on $\mathbf{X}_0 \in \mathcal{M}$. Inference can produce off-manifold $\tilde{\mathbf{X}}_0$ due to Claim 1. The impact of this drift is amplified by Claim 2.
Manifold-drift forcing robustifies the decoder by training it on interpolated latents:
\begin{equation}
\tilde{\mathcal{V}} = \alpha\,\hat{\mathcal{V}}_{0}^{(t)} + (1-\alpha)\,\mathcal{V}, \qquad t \sim \mathcal{U}([T_1, T_2]), \quad \alpha \sim \mathcal{U}([0, 1]).
\end{equation}
In flattened joint form, this first-order approximation of the drift direction is written as:
\begin{equation}
\tilde{\mathbf{X}} = (1-\alpha)\mathbf{X}_0 + \alpha\,\hat{\mathbf{X}}_0^{(t)}.
\end{equation}
This increases decoder robustness to latents produced by diffusion rollouts, forcing it to act as a projection operator back to valid 3D geometry.

\section{More Visualization}
\label{sec:more_vis}

\begin{figure}[t]
    \centering
    \includegraphics[width=0.95\linewidth]{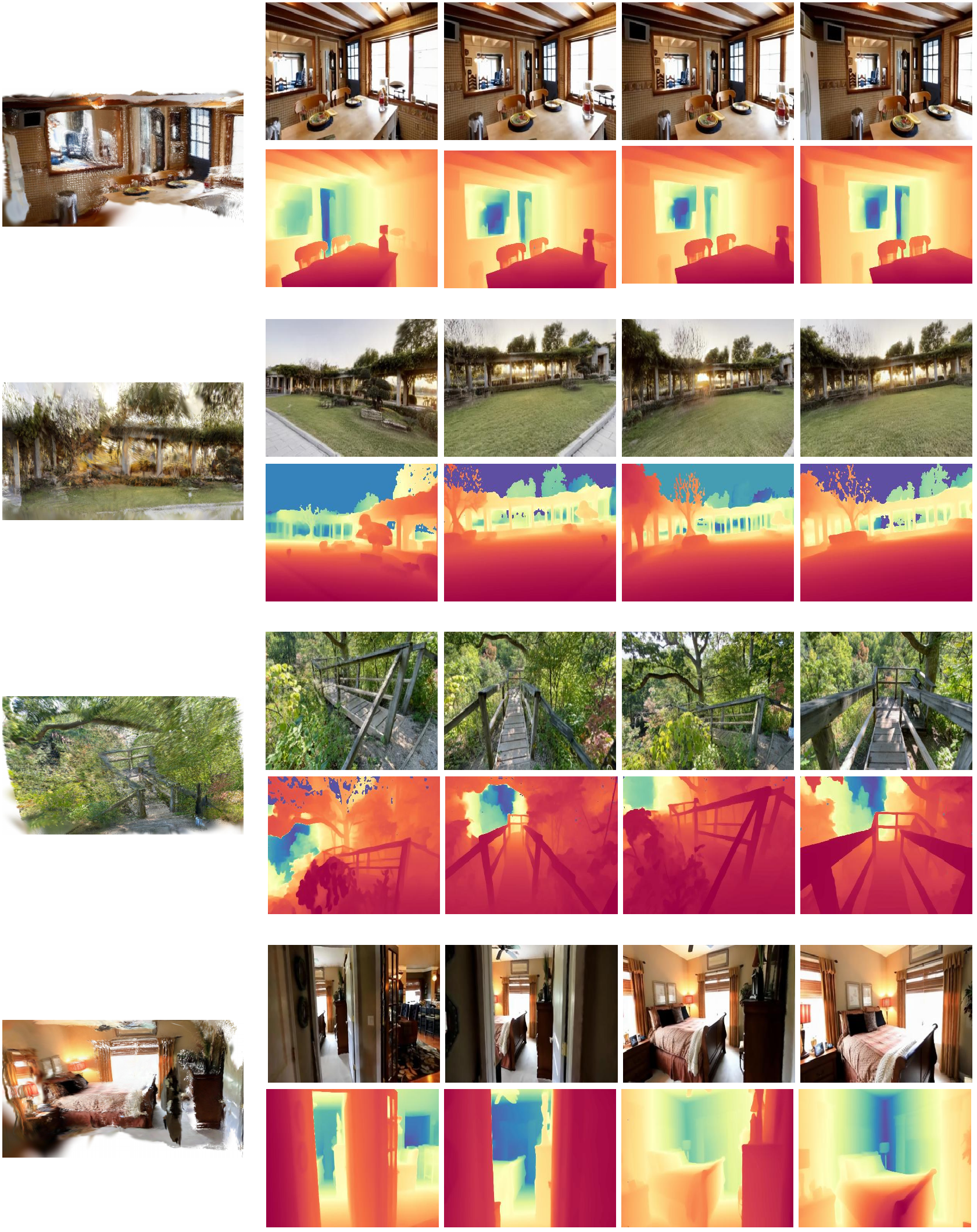}
    \caption{\textbf{Additional visualizations (I).} 3DGS structure visualizations and the corresponding novel-view RGB renderings and depth map visualizations for OneWorld-generated 3D scenes.}
    \label{fig:appendix_vis1}
\end{figure}

\begin{figure}[t]
    \centering
    \includegraphics[width=0.95\linewidth]{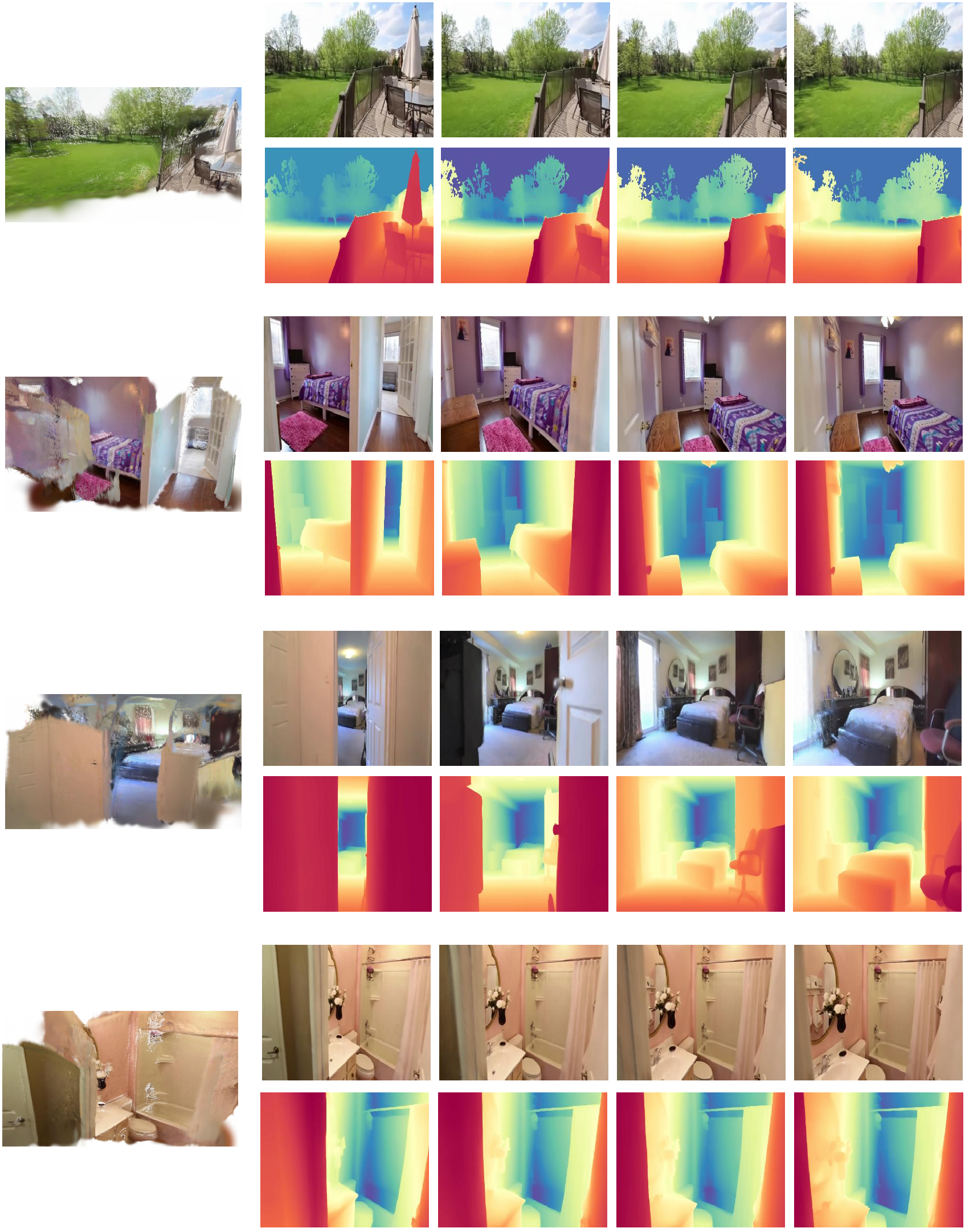}
    \caption{\textbf{Additional visualizations (II).} More 3DGS structure visualizations with novel-view RGB renderings and depth map visualizations for OneWorld-generated 3D scenes.}
    \label{fig:appendix_vis2}
\end{figure}

\begin{figure}[t]
    \centering
    \includegraphics[width=0.95\linewidth]{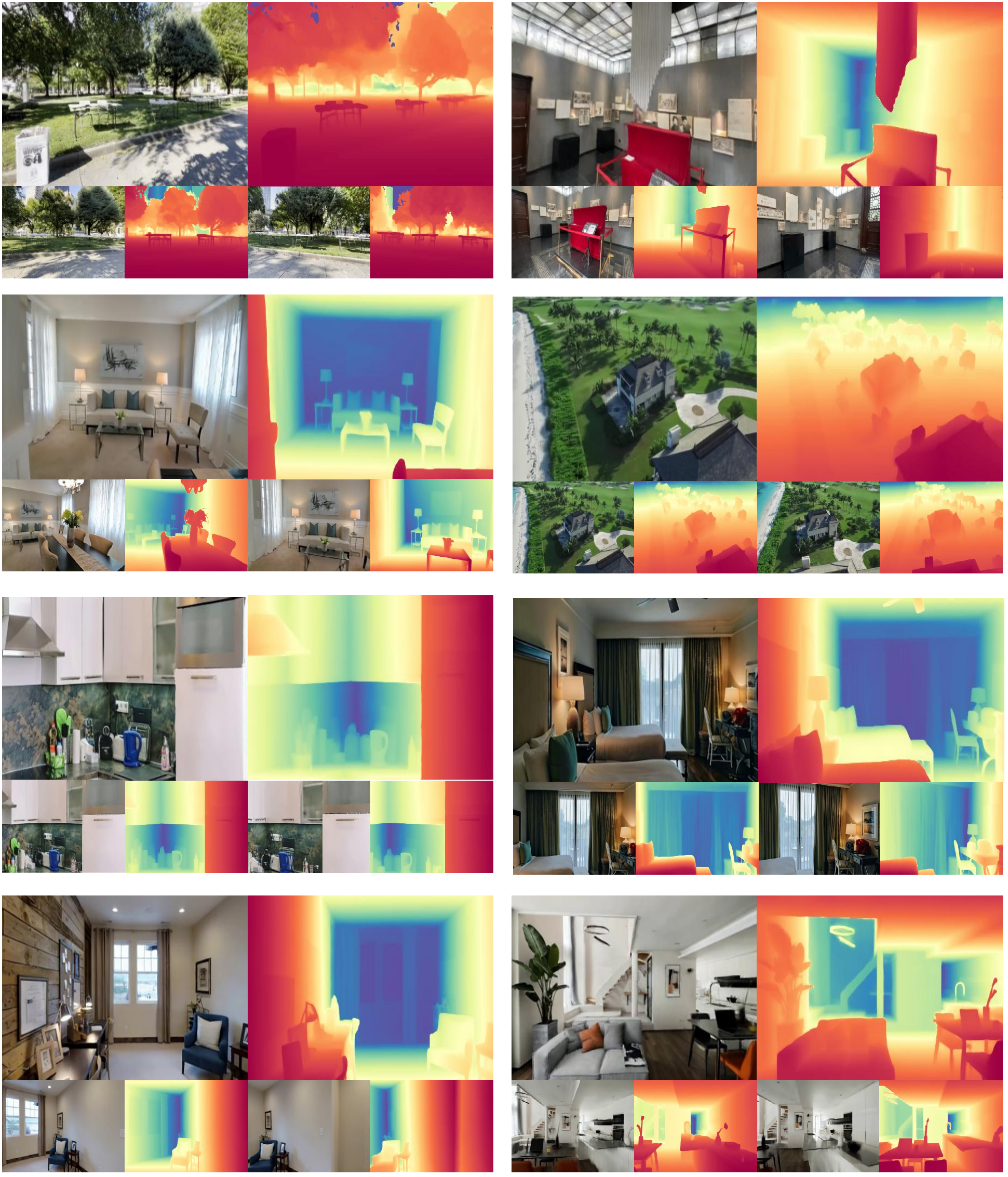}
    \caption{\textbf{More examples.} Additional 3DGS-rendered novel-view RGB images and the corresponding depth map visualizations for OneWorld-generated 3D scenes.}
    \label{fig:appendix_vis3}
\end{figure}

In this section, we provide additional visualizations for 3D scenes generated by OneWorld.
Fig.~\ref{fig:appendix_vis1} and Fig.~\ref{fig:appendix_vis2} show the 3D Gaussian Splatting (3DGS) structure visualizations, together with the corresponding novel-view RGB renderings and depth map visualizations.
Fig.~\ref{fig:appendix_vis3} further presents more examples of 3DGS-rendered novel-view RGB images and depth maps, covering diverse scenes and viewpoints.

\section{Discussion}
\label{sec:discussion}
\subsection{Limitation}
Our current model is trained on datasets that are still limited in scale and diversity, which can reduce robustness when generalizing to rare scene types, extreme viewpoints, or uncommon appearance distributions. In addition, we train and decode at a relatively low resolution, which may limit fine-grained texture fidelity and thin-structure rendering quality. In future work, we plan to scale training to larger and more diverse multi-view corpora and adopt higher-resolution training/decoding pipelines to further improve visual quality and generalization.

\subsection{LLM Usage}
AI assistants (ChatGPT) were used to correct potential grammatical inaccuracies in the manuscript. AI assistants did not participate in research ideation, experimental design, method development, result interpretation, or drawing scientific conclusions.

\clearpage
{
\bibliographystyle{plainnat}
\bibliography{main}
}

\end{document}